\pdfoutput=1
\documentclass[11pt,a4paper]{article}
\usepackage[a4paper,total={16cm,25cm}]{geometry}

\usepackage{graphicx} \usepackage{multirow} \usepackage{amsmath,amssymb,amsfonts}
\usepackage{booktabs} \usepackage{array} \usepackage{xcolor} \usepackage{textcomp}
\usepackage[numbers]{natbib}
\usepackage{appendix}
\usepackage{authblk}
\usepackage{url}
\usepackage[hidelinks]{hyperref}

\providecommand{\burl}[1]{\url{#1}}

\newcommand{\dd}{\Delta} \newcommand{\muhat}{\widehat{\mu}} \newcommand{\Sighat}{\widehat{\Sigma}} \newcommand{\Phat}{\widehat{P}}

\title{LLM knowledge for improved Out-of-Distribution Tabular Prediction -- A Cross-Validated Regularisation Approach}
\author[1]{S{\o}lve Langseth}
\author[1]{Kristoffer Herland Hellton}
\author[1,2,3]{Hugo L. Hammer\thanks{Corresponding author: \texttt{hugo.hammer@oslomet.no}}}
\affil[1]{Department of Computer Science, Oslo Metropolitan University, Oslo, Norway}
\affil[2]{Department of Holistic Systems, SimulaMet, Oslo, Norway}
\affil[3]{Department of Plastic and Reconstructive Surgery, Oslo University Hospital, Oslo, Norway}
\date{}

\begin{document}

\maketitle

\begin{abstract}
Prediction models degrade when the deployment population differs from the one used to train them, and training data cannot say which feature--response relationships will transfer. Expert elicitation is slow and unscalable. We ask whether coefficient priors elicited from large language models (LLMs) improve out-of-distribution (OOD) prediction as the shrinkage target of a regularised regression instead of zero, across five tabular datasets split on real covariates, selecting providers and prior structures by source-domain cross-validation alone. The priors help on four of five datasets, by up to 7.9\% OOD mean squared error, and ten of twenty-four prior-informed results resolve the mean improvement at conventional confidence; single-deployment intervals are much wider. On classification the gain is in probabilistic accuracy, not ranking, and control tests with corrupted priors attribute the benefit to feature-specific content rather than scale. Two findings temper this. Variation across candidate priors exceeds the average benefit, yet cross-validation margins fall below their own noise, so selection avoids the worst prior rather than finding the best. On three datasets the prior improves in-domain accuracy as much as out-of-domain, making it better regularisation than shift correction. Only where the training data are large enough that the in-domain gain disappears does elicited knowledge demonstrably improve transfer. Code, priors and results: \url{https://github.com/hugohammer/llm-priors-ood}.
\end{abstract}


\section{Introduction}

Prediction models are usually developed under the assumption that training and evaluation data are drawn from the same underlying distribution. Under this assumption, held-out test performance is a reliable proxy for the data a model will meet after deployment. In applied settings the assumption is often unrealistic, and performance can deteriorate when, for instance, the population encountered at prediction time differs from the one used for development \citep{liu2021oodsurvey}. As an example, a clinical prediction model trained at one hospital may perform worse at another because patient demographics, measurement routines, disease prevalence or treatment practices might differ.

Such changes are described as dataset or domain shift. We use \emph{domain shift} for settings where the deployment population differs from the source population used for model development, the case most relevant here; it is closely related to covariate shift, where the feature distribution changes while the conditional outcome mechanism is assumed stable \citep{quinonerocandela2009shift, morenotorres2012unifying, liu2021oodsurvey}.

Despite extensive work, there is no generally reliable remedy. Standard validation can fail precisely because the validation set is drawn from the source distribution. The dominant practical response is ensembling: deep ensembles \citep{lakshminarayanan2017deepensembles} train several networks from different initialisations and average their predictive distributions, producing better-calibrated uncertainty than more involved Bayesian methods, with ensemble variance rising on inputs far from the training distribution. Ensembles have limitations, however. They require fitting many models, are expensive, and when built from complex learners are hard to interpret. More fundamentally, ensemble disagreement diagnoses predictive uncertainty, but does not reduce it: it tells the user how confident the model is, not how to make a better prediction.

One of the few principled ways to improve predictions when training data are limited or unrepresentative is to bring in information from outside the training set: external knowledge about which features should matter, in what direction, 
and with what plausible magnitude. Recent evidence that tabular foundation models excel on small-to-medium IID data but are still beaten by conventional models on non-IID problems \citep{purucker2026beyondiid} suggests that scale alone does not solve domain shift, and that simple, inspectable models guided by external knowledge remain worth studying.

Useful external information often exists outside the training data, in previous studies, biological pathways, clinical expertise or established domain theory \citep{sokolov2016pathway, ohagan2019elicitation}. Appropriately encoded, it can improve estimation and prediction where observed data are limited, noisy or unrepresentative \citep{vandeschoot2021bayesian}. The established framework is Bayesian modelling, where prior beliefs are expressed as a distribution over parameters and updated with data \citep{bishop2006prml, vandeschoot2021bayesian}. A closely related approach is regularisation, where the estimator is penalised so that estimates shrink toward preferred values \citep{hastie2009elements}. Ridge regression shrinks coefficients toward zero, which is exactly a generic prior belief that effects are small unless the data say otherwise \citep{hoerl1970ridge, bishop2006prml}.

This equivalence makes prior-informed regularisation a natural way to combine external knowledge with predictive modelling: instead of shrinking every coefficient toward the same uninformative target of zero, one may shrink toward values suggested by domain knowledge. Under domain shift this may be helpful, because source-domain data can overemphasise relationships that do not transfer, while domain knowledge may encode more stable information about plausible effects.

Obtaining that knowledge, however, is the bottleneck. O'Hagan~\cite{ohagan2019elicitation} describes expert elicitation as a structured and demanding process of translating substantive knowledge into probability distributions.
Domain experts are usually not statisticians, such that the elicitor typically must guide them through uncertainty, probability and numerical effect sizes. Elicitation is therefore slow, hard to scale, and offers no guarantee of a useful prior.

These constraints motivate interest in Large Language Models (LLMs) as an alternative. LLMs are trained on large text corpora and adapted to respond to natural-language prompts \citep{brown2020gpt3, bommasani2021foundation}, encoding broad statistical associations across many domains \citep{bommasani2021foundation, minaee2024llmsurvey}. They can be queried repeatedly and cheaply, applied across domains and feature sets, and constrained to emit structured numerical output. In a randomised trial of diagnostic reasoning the language model alone scored higher than physicians using conventional resources, even though giving physicians access to it did not improve their own reasoning \citep{goh2024llmdiagnostic}, and LLM-derived priors have already been reported to be preferable to expert elicitation in an applied Bayesian survival analysis \citep{evans2025survival}. This does not make LLMs substitutes for human experts, but it does suggest they contain domain information worth eliciting.

The caveats are real. LLMs hallucinate, are sensitive to prompt wording, and reflect biases in their training data \citep{bender2021parrots, liu2023prompt, gallegos2024bias, huang2025hallucination}. These matter especially when the output is a numerical coefficient, since errors in sign, scale or uncertainty propagate directly into the fitted model.

We address three questions. First, can prior means estimated from repeated LLM coefficient samples improve OOD performance for regularised linear and logistic regression, relative to standard shrinkage toward zero? Second, does estimating prior uncertainty from the empirical distribution of LLM samples (as feature-specific variances or full covariance) help beyond the prior mean alone? Third, under what conditions do LLM priors help, do nothing, or harm?

Our contributions are:

\begin{enumerate}\itemsep2pt
\item An evaluation of LLM-elicited coefficient priors under domain splits
  defined by real covariates across five datasets, with a cross-provider,
  cross-family selection pipeline driven only by source-domain data.
\item Placebo priors: priors re-indexed onto permuted features, and priors
  with randomly flipped signs, both preserving scale and correlation
  structure, establishing that the benefit derives from the priors'
  feature-specific content and not from shrinking toward an arbitrary non-zero
  target.
\item A methodological negative result: Comparing classifiers using fixed decision thresholds (e.g., computing $F_1$ at $0.5$) substantially overstates performance gains whenever the domain split alters class balance. The reliable benefit of prior-informed regularisation lies in probabilistic calibration and proper scoring rules rather than threshold-dependent metrics~\citep{gneiting2007scoring}.
\item A separation of two aspects of the benefit: on three of five
  datasets the prior improves in-domain accuracy as much as out-of-domain
  accuracy and is therefore better regularisation, while on the one dataset
  with enough training data for the in-domain gain to disappear it improves transfer specifically.
\item An assessment of the selection step itself: depending on the LLM variant and prior family, an elicited prior can either improve or degrade out-of-domain predictive performance relative to the zero-centred baseline. Yet the cross-validation margins separating candidates are smaller than their fold-level noise, so the selection procedure is better at avoiding poor candidates than at identifying the best one.
\end{enumerate}

\section{Related work}

The use of LLMs as a source of domain information for predictive modelling is recent but developing quickly. Existing work uses LLMs directly as predictors, to construct or modify features, or to supply prior information for statistical models.

This study builds on Riegler et al.~\cite{riegler2025priors}, who asked several LLMs to suggest informative Gaussian priors for coefficients in linear and logistic regression. They found that LLMs often identify the correct \emph{direction} of a feature effect but calibrate prior \emph{width} poorly, exhibiting both over- and under-confidence depending on the variable and the provider. That finding motivates our estimator design directly, and Riegler et al.~\cite{riegler2025priors} explicitly name out-of-distribution (OOD) prediction as a future use case, which is the focus of this paper.

Capstick et al.~\cite{capstick2025autoelicit} propose AutoElicit, which asks an LLM for its best guess of the mean and standard deviation of each coefficient under $K = 100$ automatically rephrased task descriptions, combining the results into a mixture prior $\sum_{k} \pi_{k} \mathcal{N}(\theta \mid \mu_{k}, \sigma^{2}_{k})$, for Bayesian linear and logistic regression, showing that LLM priors can beat uninformative priors and reduce labelling effort substantially. Its evaluation is in-distribution, and asks how few labels a well-specified prior saves rather than how well the fitted model transfers. Huang~\cite{huang2025llmprior} generalises the elicitation target, converting unstructured context into tractable distributions via a Gaussian mixture and aggregating priors across agents by logarithmic opinion pooling. Zhang et al.~\cite{zhang2025llmlasso} take a different encoding approach with LLM-Lasso, asking the LLM for feature-specific penalty scores that guide Lasso regularisation rather than for coefficient values.

Other work bypasses coefficients entirely: Requeima et al.~\cite{requeima2024llmprocesses} treat the LLM as an in-context probabilistic regressor competitive with Gaussian processes, and Zeng et al.~\cite{zeng2024llmembeddingsimprovetesttime} train on LLM embeddings of serialised rows, finding inconsistent robustness under $(Y \mid X)$ shift unless a few labelled target examples are available. Amin et al.~\cite{amin2026integratingmachinelearningensembles} find that an ML ensemble substantially outperforms LLM prediction alone, with hybrid fusion adding only a small gain, supporting the broader principle that LLMs are more useful as structured inputs to classical models than as final predictors.

AutoElicit is the method closest to our own. While we detail their exact relationship below, their primary focus is in-distribution prediction, whereas ours is OOD generalisation. By collapsing AutoElicit's mixture to a single Gaussian through moment matching, we obtain a diagonal Gaussian prior over the coefficients. This is the same object as our suggested Independent-Prior of Table~\ref{tab:families}. Three characteristics then separate the two methodologies. The first is where the prior's width originates: AutoElicit asks the model to state a standard deviation, whereas we take the empirical spread of $M$ sampled coefficient vectors, so its widths are self-reported and ours are behavioural. The second is the source of diversity, $K$ rephrasing of the task description against $M$ samples at a fixed prompt, which capture cross-prompt and within-prompt variability respectively and are complementary rather than competing. The third is the one our results make central: AutoElicit uses the elicited prior at the strength the model states, with no free parameter governing how far to trust it, whereas we introduce $\lambda$ and let source-domain cross-validation rescale the entire precision. 

Closest to this work is LoID \citep{rezaeimanesh2026languagemodelsknowdont}, which extracts priors for Bayesian logistic regression from token-level predictions rather than generated coefficients, comparing the model's preference for semantically opposed tokens across prompt phrasings and using the variation across phrasings as uncertainty. LoID is evaluated on fifteen tabular datasets, fourteen public and one private, under OOD covariate shift, and outperforms LLM Processes on thirteen of them and AutoElicit on eleven. We differ in the elicitation mechanism (repeated coefficient sampling rather than token log-probabilities), in the encoding of prior uncertainty (full covariance across coefficients, not only independent variances), and in how absolute prior strength is set (source-domain cross-validation rather than fixed hyperparameters). We also differ in how the shift itself is constructed. LoID restricts its training set to a region of a real covariate (excluding respondents under $28$ on Adult, or keeping only normoglycaemic patients on Diabetes) and evaluates on a held-out sample of the undivided population, so the target contains the training region as well as the part withheld from it. We train on one domain and test on a disjoint other, and on two datasets the division is not ours to choose at all: Heart Disease splits by recruiting cohort and Cancer by geographic region.

\section{Methods}

\subsection{Notation and problem setting}\label{subsec:notation}

Let $x \in \mathbb{R}^{p}$ denote a column vector of $p$ features and $y$ the associated response, real-valued for regression and binary for classification. A \emph{domain} is a joint distribution over $(x, y)$. We write $P_{S}$ for the source domain, from which the available data are drawn, and $P_{T}$ for the target or deployment domain, on which the fitted model will actually be used. The training sample is $\mathcal{D}_{n} = \{(x_{i}, y_{i})\}_{i=1}^{n}$, drawn independently from $P_{S}$.

All models in this study are linear in the features, with linear predictor
\begin{equation}
\eta_{\beta}(x) \;=\; \beta_{0} + x^{\top}\beta ,
\qquad \beta \in \mathbb{R}^{p},\ \beta_{0} \in \mathbb{R},
\label{eq:linpred}
\end{equation}
where $\beta$ is the coefficient vector and $\beta_{0}$ an intercept. For regression we assume $y = \eta_{\beta}(x) + \varepsilon$ with $\varepsilon \sim \mathcal{N}(0, \sigma^{2})$; for classification, $\Pr(y = 1 \mid x) = \mathrm{logit}^{-1}(\eta_{\beta}(x))$. In both cases we write $\mathcal{L}(\beta; \mathcal{D}_{n})$ for the negative log-likelihood of the sample, which reduces to the residual sum of squares in the regression case and to the Bernoulli negative log-likelihood in the classification case.

The quantity we would like to make small is the risk on the deployment domain,
\begin{equation}
R_{T}(\beta) \;=\; \mathbb{E}_{(x,y) \sim P_{T}}\bigl[\ell\bigl(y, \eta_{\beta}(x)\bigr)\bigr],
\label{eq:risk}
\end{equation}
for a loss $\ell$. Standard estimation instead minimises an empirical version of the \emph{source} risk $R_{S}$, and the two coincide only when $P_{S} = P_{T}$.

When they differ we speak of domain shift, and it is worth distinguishing three kinds \cite{quinonerocandela2009shift, morenotorres2012unifying}. Under \emph{covariate shift} the feature distribution changes, $P_{S}(x) \neq P_{T}(x)$, while the conditional mechanism $P(y \mid x)$ is unchanged. Under \emph{label shift} the marginal response distribution changes, $P_{S}(y) \neq P_{T}(y)$, for instance when the target population contains a larger proportion of cases than the source. Under \emph{concept shift} the association itself changes, $P_{S}(y \mid x) \neq P_{T}(y \mid x)$, as when the risk conferred by a given exposure differs between populations.

The distinction matters for what may be assumed rather than for what we do. The splits used here are defined by real covariates: housing stock age, census region, clinical cohort, patient age, employment sector, and such splits generally induce more than one kind of shift at once. In particular we have no way to verify that $P(y \mid x)$ is preserved across, say, American and European cardiology cohorts, so concept shift is a live possibility on several of our datasets and we do not assume it away. This is a reason for the approach rather than an obstacle to it: if the conditional mechanism itself differs between domains, no reweighting of the source sample can recover the target relationship, and external knowledge about which effects are stable becomes the only available lever. Label shift matters separately and concretely, because it is what makes a fixed decision threshold inappropriate on the target domain, a point we return to in the Results.

The difficulty is one of identification rather than of estimation. Write $\beta^{\star}_{S}$ and $\beta^{\star}_{T}$ for the coefficient vectors minimising the source and target risks. Maximum likelihood on $\mathcal{D}_{n}$ is consistent for $\beta^{\star}_{S}$, and no procedure that sees only $\mathcal{D}_{n}$ can separate the components of $\beta^{\star}_{S}$ that also hold under $P_{T}$ from those that are artefacts of the source population, because $P_{T}$ is simply not represented in the sample. Cross-validation does not help: its folds are themselves draws from $P_{S}$. Improving $R_{T}$ therefore requires information from outside the training sample.

We refer to such information as \emph{global knowledge}: statements about the relationship between features and response that are not derived from either sample, and that are expected to hold across populations rather than within one. Established domain theory, mechanistic understanding, results from previous studies on different populations, and the judgement of experienced practitioners are all of this kind.

Such knowledge enters the model as a distribution over the coefficients. Write $f(\beta)$ for a density on $\mathbb{R}^{p}$ expressing what is believed about $\beta$ before the sample is seen; in the Bayesian literature this is the \emph{prior} distribution. The sample enters through the \emph{likelihood} $f(\mathcal{D}_{n} \mid \beta) = \exp\{-\mathcal{L}(\beta; \mathcal{D}_{n})\}$, and the two combine into the \emph{posterior} distribution
\begin{equation}
f(\beta \mid \mathcal{D}_{n}) \;\propto\; f(\mathcal{D}_{n} \mid \beta)\, f(\beta).
\label{eq:posterior}
\end{equation}
The two factors pull in different directions, and that is the point. The likelihood is a statement about the source sample alone and is maximised at $\beta^{\star}_{S}$, whereas $f(\beta)$ does not depend on that sample at all. Multiplying them moves the posterior off the source optimum and toward wherever the prior places its mass, by an amount governed by how concentrated the prior is. If the prior encodes global knowledge, this is exactly the correction the source sample cannot supply on its own. The notation $f(\beta)$, $f(\mathcal{D}_{n} \mid \beta)$ and $f(\beta \mid \mathcal{D}_{n})$ is used throughout the rest of this section.

\subsection{Penalised regression as maximum a posteriori estimation}

The standard device for controlling variance in a linear model is an $L_{2}$ penalty,
\begin{equation}
\widehat{\beta}_{\text{ridge}} \;=\; \arg\min_{\beta}\;
\Bigl\{ \mathcal{L}(\beta; \mathcal{D}_{n}) + \tfrac{1}{2}\lambda \lVert \beta \rVert_{2}^{2} \Bigr\},
\qquad \lambda \ge 0,
\label{eq:ridge}
\end{equation}
which gives ridge regression in the Gaussian case and $L_{2}$-regularised logistic regression in the binary case \cite{hoerl1970ridge, hastie2009elements}. Penalisation trades a small increase in bias for a larger reduction in variance, and is especially useful when $p$ is large relative to $n$ or the features are collinear.

Equation~Eq.~\ref{eq:ridge} has an exact Bayesian reading. Placing the prior $\beta \sim \mathcal{N}(0, \lambda^{-1} I_{p})$ on the coefficients and combining it with the likelihood gives a posterior whose mode is precisely $\widehat{\beta}_{\text{ridge}}$, since the negative log-prior equals $\tfrac{1}{2}\lambda\lVert\beta\rVert_{2}^{2}$ up to an additive constant \cite[§3.4]{hastie2009elements}, \cite{bishop2006prml}. Ridge regression is therefore maximum a posteriori (MAP) estimation under a particular prior, and $\lambda$ is the prior precision: large $\lambda$ expresses a tight prior concentrated at the origin, and $\lambda \to 0$ recovers the unpenalised estimator.

This reading makes the content of the ridge penalty explicit, and it is worth stating plainly: ridge asserts that every coefficient is \emph{a priori} centred at zero. That is a statement of ignorance, not of knowledge. It is a reasonable default and it does reduce variance, which is valuable under domain shift because a model that has fitted source-specific structure will generalise poorly. But shrinking toward zero provides no information about \emph{which} effects transfer between populations, and it is exactly that information which the source sample lacks.

More generally, penalisation pulls the estimate toward the target, which need not mean making it smaller: if the target lies further from the origin than the unpenalised estimate does, the penalty moves the coefficients \emph{outward} and $\lVert\beta\rVert$ grows with $\lambda$. ``Shrinkage'' is then a misnomer for what the penalty does to the norm, although it remains accurate about what it does to the variance, which falls either way because the estimate is held close to a fixed point instead of varying freely with the sample.
It reduces bias with respect to $R_{T}$ only if the target lies closer to $\beta^{\star}_{T}$ than the unshrunk estimate does. The origin is one possible target; global knowledge suggests others.

\subsection{Priors that encode external knowledge}

Replacing the zero-mean isotropic prior with a general Gaussian $\beta \sim \mathcal{N}(\mu, \lambda^{-1} P^{-1})$, where $\mu \in \mathbb{R}^{p}$ and $P$ is positive definite, and taking the posterior mode gives
\begin{equation}
\widehat{\beta}(\mu, P, \lambda) \;=\; \arg\min_{\beta}\;
\Bigl\{ \mathcal{L}(\beta; \mathcal{D}_{n})
+ \tfrac{1}{2}\lambda\,(\beta - \mu)^{\top} P\,(\beta - \mu) \Bigr\}.
\label{eq:map}
\end{equation}
Here $\mu$ is the shrinkage \emph{target} and $P$ the shrinkage \emph{structure}; $\lambda$ scales the whole penalty. Ridge is the special case $(\mu, P) = (0, I_{p})$. The intercept is never penalised, as it is determined by the response base rate and the centring of the features rather than by any belief about feature--response relationships.

Three choices of $P$ correspond to three strengths of statement that a source of knowledge might make:
\begin{enumerate}\itemsep2pt
\item $P = I_{p}$ uses only a point belief: every coefficient is pulled toward its target $\mu_{j}$ with the same strength.
\item $P = \mathrm{diag}(1/\sigma_{1}^{2}, \dots, 1/\sigma_{p}^{2})$ adds per-coefficient confidence. Coefficients the source is sure about have small $\sigma_{j}^{2}$ and are held near $\mu_{j}$; coefficients it is unsure about are left to the data.
\item $P = \Sigma^{-1}$ for a full covariance $\Sigma$ additionally encodes dependence between beliefs, allowing the prior to express that two coefficients are believed to move together even when neither is determined precisely on its own.
\end{enumerate}

These quantities have a natural interpretation when knowledge is elicited from several sources. Suppose $M$ experts are each asked for a complete coefficient vector, yielding $\beta^{(1)}, \dots, \beta^{(M)} \in \mathbb{R}^{p}$. The pooled belief and the disagreement among experts are then estimated by
\begin{equation}
\begin{aligned}
\widehat{\mu} &= \frac{1}{M}\sum_{m=1}^{M} \beta^{(m)},
\qquad
\widehat{\sigma}^{2}_{j} = \frac{1}{M-1}\sum_{m=1}^{M}
   \bigl(\beta^{(m)}_{j} - \widehat{\mu}_{j}\bigr)^{2}, \\[4pt]
\widehat{\Sigma} &= \frac{1}{M-1}\sum_{m=1}^{M}
   \bigl(\beta^{(m)} - \widehat{\mu}\bigr)\bigl(\beta^{(m)} - \widehat{\mu}\bigr)^{\top}.
\end{aligned}
\label{eq:priorquant}
\end{equation}
Disagreement between experts becomes prior uncertainty, which is the classical aggregation view of structured elicitation \cite{ohagan2019elicitation, vandeschoot2021bayesian}. Nothing in Eq.~\ref{eq:priorquant} depends on what the experts are.

Moving the target away from the origin has a cost that is easy to overlook. With $\mu = 0$, increasing $\lambda$ shrinks $\lVert\beta\rVert$ and therefore reduces variance monotonically. With $\mu \neq 0$, increasing $\lambda$ drives $\widehat{\beta}$ toward $\mu$ rather than toward zero, so the implicit norm control that gives ridge its stability is given up: if $\mu$ is wrong, a large $\lambda$ now moves the estimate away from the target optimum. The parameter $\lambda$ thus governs a balance between two distinct things: how much the data are trusted relative to the prior, and how much variance reduction is retained, and the balance is favourable only to the extent that $\mu$ is informative.

One way to hedge against a misspecified target without abandoning it is to combine the two extremes at the level of predictions rather than coefficients. Writing $g$ for the inverse link (the identity for regression and $\mathrm{logit}^{-1}$ for classification) the blend is taken on the response scale,
\begin{equation}
\widehat{y}_{\gamma}(x) \;=\; (1 - \gamma)\,g\!\left(\eta_{\widehat{\beta}(0, I, \lambda_{0})}(x)\right)
\;+\; \gamma\,g\!\left(\eta_{\widehat{\beta}(\mu, I, \lambda_{1})}(x)\right),
\qquad \gamma \in [0, 1],
\label{eq:mixed}
\end{equation}
so that for classification it is the two predicted probabilities that are averaged, not the two sets of log-odds. The distinction is immaterial for regression, where $g$ is the identity and the two coincide, but it defines a different model in the logistic case, and averaging probabilities is what the fitted results below use.

which recovers the uninformative baseline at $\gamma = 0$ and the fully informed model at $\gamma = 1$. 

\subsection{Selecting prior hyperparameters using LLMs}

The prior in Eq.~\ref{eq:map} is itself parameterised, by the quantities $(\mu, P, \lambda)$; in the Bayesian literature these are the prior \emph{hyperparameters}, and they must be given values before the model can be fitted. Placing further distributions on them only defers the choice, since those distributions have parameters of their own, so at some level values have to be set directly \cite{vandeschoot2021bayesian}. We set them at this first level, which keeps the estimator comparable to standard penalised regression and lets one implementation cover both the uninformative and the informative case. What remains is how each of them is obtained, and the three are not on the same footing: it is useful to separate the \emph{shape} of the prior, $(\mu, P)$, from its overall \emph{scale}, $\lambda$. The distinction matters because elicitation is generally far more reliable about relative statements: which effects are positive, which are larger than others, which are more uncertain, than about the absolute magnitude of a coefficient or the width of a prior interval \cite{ohagan2019elicitation}. The same asymmetry has been reported specifically for language-model elicitation, where the direction of an effect is usually right while the stated width is poorly calibrated in both directions \cite{riegler2025priors}. Committing to an elicited scale would therefore import the least reliable part of the judgement.

We consequently take the shape from elicitation and estimate the scale from data, choosing $\lambda$ by cross-validation on the source sample. The division of labour is deliberate: the prior supplies the information it is reliable about, and the data decide how far to trust it. It also means the procedure degrades gracefully, since an uninformative or actively misleading prior can be switched off by driving $\lambda$ toward zero.

Everything in this section is independent of where the vectors $\beta^{(1)}, \dots, \beta^{(M)}$ come from. Obtaining them from human experts is the classical approach, and it is slow, costly and hard to scale. The central question of this paper \textit{is whether LLMs can supply them instead}, cheaply and at scale, and whether the resulting priors improve prediction on a shifted deployment domain. The remainder of this section describes how we elicit priors from LLMs and how we evaluate the resulting predictive performance on domain splits defined by real covariates. 

\subsection{Pipeline overview}

\begin{figure}[htbp]
\centering
\includegraphics[width=\linewidth]{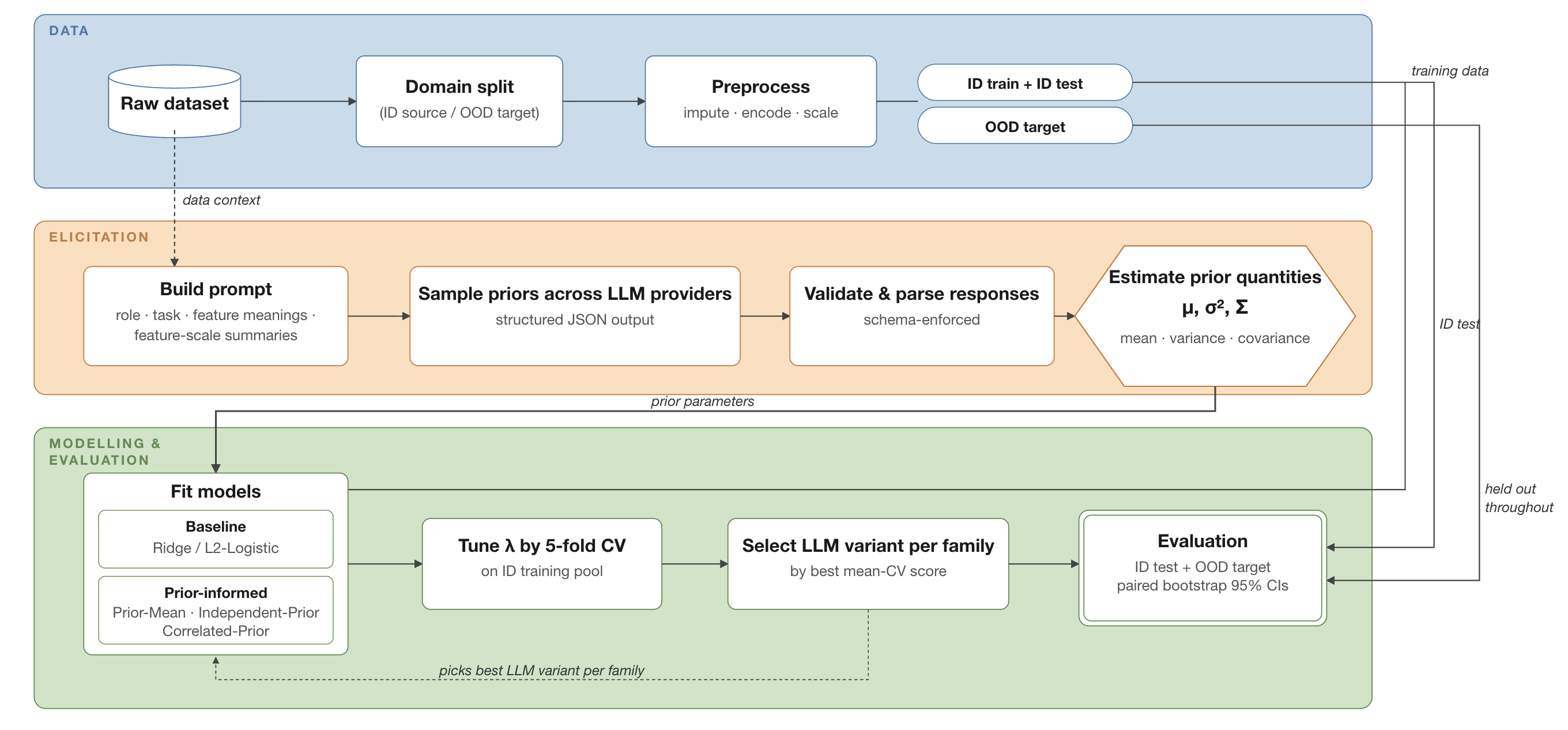}
\caption{Pipeline overview. The data branch (top) fixes the ID/OOD split and fits preprocessing on source data only. The elicitation branch (middle) feeds dataset context into prompt construction, draws repeated samples per provider, validates them against a JSON schema, and estimates $\muhat$, $\widehat{\sigma}^{2}$ and $\Sighat$ as in Eq.~\ref{eq:priorquant}. The modelling branch (bottom) fits each family, tunes $\lambda$ by 5-fold CV on the source training pool, selects the winning variant per family by mean CV score, and evaluates on the held-out ID test and OOD sets. The OOD set never enters any selection step.}
\label{fig:pipeline}
\end{figure}

Figure~\ref{fig:pipeline} shows how the steps fit together. Three branches meet at the modelling step. The data branch fixes an OOD partition up front and fits all preprocessing on the source domain alone. 
The elicitation branch queries several language models repeatedly for coefficient vectors and converts the resulting samples into the prior quantities of Eq.~\ref{eq:priorquant}. The modelling branch solves Eq.~\ref{eq:map} for each choice of $(\mu, P)$, tunes $\lambda$, and chooses among providers and prior structures.

Each dataset is divided into a source partition and a held-out partition drawn from a systematically different population. The held-out partition plays the role of the target domain $P_{T}$ of Section~\ref{subsec:notation}: it is the sample on which $R_{T}$ is estimated. Following the convention in this literature we call it the out-of-distribution (OOD) partition and the source side in-domain (ID), but the two names denote exactly the two domains already introduced.

One principle governs the whole design and is worth stating before the detail: \emph{no quantity derived from the OOD partition enters any fitting, tuning or selection step}. Preprocessing statistics, the prior strength $\lambda$, the winning language-model variant and the prior family are all determined on source-domain data only, which is the situation a practitioner is actually in. The OOD set is touched once, at evaluation.

The model families used in the experiments are instances of Eq.~\ref{eq:map} and differ only in the
pair $(\mu, P)$; Table~\ref{tab:families} collects them. Reading down the table, the prior is given
progressively more structure to express: nothing at all beyond a preference for small coefficients,
then a target, then a target with per-coefficient confidence, then a target with a full covariance.
The number of elicited quantities entering the penalty therefore grows from none, through $p$, to
$p + p$, to $p + p(p+1)/2$, which is also the order in which the estimators become harder to
estimate reliably from a finite number of elicitations. The last two rows are not points on that
scale: Mixed hedges between the first two by blending their predictions, and the unregularised fit
is the $\lambda = 0$ limit in which the prior drops out entirely and is reported as a reference.

\begin{table}[htbp]
\centering
\footnotesize
\caption{The fitted families as choices of $(\mu, P)$ in Eq.~\ref{eq:map}. The intercept is never
penalised. $\widehat{\mu}$, $\widehat{\sigma}^{2}_{j}$ and $\widehat{\Sigma}$ are the elicited
quantities of Eq.~\ref{eq:priorquant}; ``elicited quantities'' counts how many numbers from the
elicitation enter the penalty.}
\label{tab:families}
\begin{tabular}{@{}llll@{}}
\toprule
Family & $\mu$ & $P$ & Elicited quantities \\
\midrule
Baseline (ridge, $L_{2}$-logistic) & $0$ & $I_{p}$ & --  \\
Prior-Mean & $\widehat{\mu}$ & $I_{p}$ & $p$ \\
Independent-Prior & $\widehat{\mu}$ & $\mathrm{diag}(1/\widehat{\sigma}^{2}_{j})$ & $2p$ \\
Correlated-Prior & $\widehat{\mu}$ & $\widehat{\Sigma}^{-1}$ & $p + p(p{+}1)/2$ \\
\midrule
Mixed & \multicolumn{2}{l}{blend of the two rows above, Eq.~Eq.~\ref{eq:mixed}} & $p$ \\
Unregularised ref. & \multicolumn{2}{l}{$\lambda = 0$; the prior drops out} & --  \\
\bottomrule
\end{tabular}
\end{table}

Two numerical details matter in practice. The elicited variances enter the Independent-Prior
precision as $1/\widehat{\sigma}^{2}_{j}$, which is unbounded if every elicitation returns the same
value for feature $j$. For the Correlated-Prior, the corresponding $\widehat{\Sigma}$ is then singular and $\widehat{\Sigma}^{-1}$ undefined. Exact agreement is not hypothetical, since a confident elicitor
asked repeatedly for the same coefficient may well repeat itself. We therefore add a small constant
to the elicited variances before inverting them, $\widehat{\sigma}^{2}_{j} + \varepsilon$ with
$\varepsilon = 10^{-8}$, and a matching jitter of $10^{-6}$ to the diagonal of
$\widehat{\Sigma}$ before inversion. These changes are not innocuous. When a unanimous
elicitation gives $\widehat{\sigma}^{2}_{j} = 0$ exactly, the
Independent-Prior penalty weight becomes $\lambda/\varepsilon = \lambda \times
10^{8}$, which fixates that coefficient at $\widehat{\mu}_{j}$ rather than merely
shrinking toward it. Far from capping the confidence a unanimous elicitation
may express, $\varepsilon$ converts it into an equality constraint, and it is a
free parameter of the estimator rather than a numerical detail. This matters
because one deployed variant does return bit-identical draws, for four of
eleven features on Boston Housing and two of seven on Diabetes, so we treat
$\varepsilon$ as something to be checked rather than asserted and report a
sensitivity analysis over it in Appendix~\ref{secA:epsilon}.

\subsection{Datasets and domain splits}

We use five publicly available tabular datasets spanning housing economics, cancer epidemiology, cardiology, diabetes screening and census income: Boston Housing~\cite{harrison1978hedonic}, a county-level cancer mortality compilation~\cite{rippner2017cancer}, the multi-cohort UCI Heart Disease database~\cite{detrano1989heart}, the Pima Indians Diabetes Database~\cite{smith1988pima} and UCI Adult~\cite{kohavi1996adult}. Two are regression tasks and three are binary classification. They were chosen to vary in the two properties we expect to govern whether a prior helps: training-set size, which ranges from $242$ to about $27{,}000$ rows, and the magnitude of the domain shift. Table~\ref{tab:datasets} summarises them.

Crucially, the domain shift in each case is defined by a real covariate rather than by a synthetic perturbation: housing stock age, US census region, clinical cohort and continent, patient age, and employment sector. Each split therefore reproduces a shift a practitioner could plausibly meet on deployment, at the cost of giving one shift per dataset rather than a controllable family of them.

One feature is removed on ethical grounds rather than statistical ones. Boston Housing's $B = 1000(\mathrm{Bk} - 0.63)^2$, with $\mathrm{Bk}$ the proportion of Black residents, encodes a racial-segregation hypothesis in its functional form and is non-invertible; it is the reason the dataset was removed from major software distributions~\cite{pedregosa2011sklearn}. We drop the column from the design matrix and the corresponding coordinate from $\muhat$ and $\Sighat$, leaving eleven features. Taking the principal submatrix of $\Sighat$ is the marginal prior over the features actually fitted. No other dataset is affected. Dropping it costs about a third of the effect on this dataset: the best resplit average falls from $+11.5\%$ to $+7.9\%$, and the Prior-Mean advantage disappears altogether, which is itself informative about how much of the original lift depended on that one covariate. The prompts still described the twelve-feature problem, so the elicited coefficients are for a specification we no longer fit; this is a mild omitted-variable mismatch that we prefer to the alternative.

Every dataset follows the same protocol. A domain-defining variable is chosen, the data are divided into in-domain (ID) source and OOD sets on that variable, and the splitting variable is excluded from the feature matrix, such that domain identity cannot leak into the model. The ID set is split $80/20$ into a training pool and a held-out ID test set; only the training pool is used for tuning and fitting. Features are dropped when they are unusable in either domain and the remainder are median-imputed from the source domain, every fill value coming from the source half. The only dataset where features are dropped is Heart Disease, where \texttt{ca}, \texttt{thal} and \texttt{slope} are discarded because they have $98\%$, $76\%$ and $50\%$ missing values in the Hungarian and Swiss cohorts, though being near-complete in Cleveland. The criterion is therefore the recording practices of the target cohorts rather than source-domain missingness alone, and we state it that way. A fourth feature, \texttt{fbs}, is missing in about $20\%$ of the target rows and was retained after an imputation audit; we report this for transparency, and repeating the whole selection without it changes the Heart Disease log-loss lift from $+8.03\%$ to $+8.00\%$, so nothing in the result depends on it. Numeric features are standardised to zero mean and unit variance, and categorical features are one-hot expanded with levels aligned across source and OOD. The imputer and scaler are fitted on the training portion only (inside each cross-validation fold during tuning, and on the full training pool at deployment) and then applied to held-out data.

\begin{table}[htbp]
\centering
\footnotesize
\caption{Datasets, domain splits and resulting partitions. ``Shifted'' counts
features with $|$Cohen's $d| > 0.5$ between source and OOD.}
\label{tab:datasets}
\begin{tabular}{@{}llrrrl@{}}
\toprule
Dataset & Domain split (source $\rightarrow$ OOD) & $p$ & $n_{\text{tr}}$ & $n_{\text{OOD}}$ & Shifted \\
\midrule
\multicolumn{6}{@{}l}{\emph{Regression}}\\
Boston Housing & housing stock: newer $\rightarrow$ older & 11 & 284 & 151 & 9/11 \\
Cancer         & US region: South $\rightarrow$ non-South & 13 & 1{,}106 & 1{,}664 & 5/13 \\
\multicolumn{6}{@{}l}{\emph{Classification}}\\
Heart Disease  & cohort: Cleveland $\rightarrow$ Hungary/Switz. & 18 & 242 & 417 & 2/18 \\
Diabetes       & patient age: $\le 29 \rightarrow\ > 29$ & 7 & 316 & 372 & 2/7 \\
Adult          & sector: private $\rightarrow$ non-private & 20 & 27{,}124 & 14{,}936 & 2/20 \\
\bottomrule
\end{tabular}
\end{table}

A sixth dataset, a small \emph{in vitro} ruminant fermentation experiment with $n_{\text{train}} = 124$ \citep{wang2017oilseeds}, was run through the same pipeline but is not reported: its per-fold validation curve was essentially flat across the whole $\lambda$ grid, the selection rule landed at $\lambda \approx 0$ for every family, and the deployed estimators either tied the unregularised fit or were marginally worse. This is a practical lower bound on when the methodology can deploy a meaningful prior: cross-validated prior-informed estimation needs a validation surface with enough signal to distinguish $\lambda$ values.

\subsection{Eliciting priors from LLMs}\label{subsec:elicit}

Each prompt has a system part containing stable elicitation instructions and a user part containing dataset-specific content. The system prompt assigns a domain-appropriate role (for example epidemiologist, cardiologist, demographic specialist), requires coefficients in raw feature units, excludes the intercept (which is determined by the outcome base rate and feature centering rather than by domain expertise), and requires conformance to a fixed JSON schema. For repeated sampling the system prompt additionally asks for one plausible draw from the model's belief distribution rather than a collapsed average, and to favour effects likely to persist under domain shift.

The user prompt supplies five things: the modelling task, domain context describing the source population, feature meanings with explicit descriptions, a short block of domain hints written by us, and quantitative feature-scale summaries (median, interquartile range and central $95\%$ range) computed from the source training pool only. These let the model reason about plausible magnitudes rather than guessing from feature names. The hint block states the expected sign of the dominant coefficients and, on some datasets, their relative size: on Diabetes, for instance, that glucose is the strongest single predictor and that BMI should be positive but smaller. It is reproduced in full for one dataset in Appendix~\ref{secA:prompt}. This is a real constraint on the interpretation of what follows: the directional content of the prior is partly ours, while the magnitudes, the per-feature variances and the covariance structure are the model's. Section~\ref{subsec:selection} and the Limitations return to what can and cannot be attributed to the elicitation as a result. Deliberately, prompts describe the source domain, but never reveal the OOD target population: the Cancer prompt states that the source consists of US South-region counties without disclosing that evaluation uses non-Southern counties. Any OOD improvement therefore cannot reflect targeted optimisation for a known shift. Because models are fitted on standardised features, raw priors are converted to standardised space by multiplying each coefficient by the corresponding feature standard deviation, recomputed per fold.

JSON output is constrained both by prompt instruction and by an API-level schema (OpenAI \texttt{json\_schema} strict mode, Anthropic tool input schema with forced \texttt{tool\_choice}, Google \texttt{response\_schema}). The exception is Anthropic extended-thinking mode, where the API forbids combining extended thinking with forced tool use; there the schema is enforced by prompt only and the parser recovers JSON from the text.

The twelve variants are six models from three providers, each queried in a
non-reasoning and a reasoning mode. Table~\ref{tab:variants} lists them with
the identifiers under which their elicited samples are stored in the
repository, which is how a reader locates the $50$ draws behind any dataset and
variant. They
are not twelve independent sources: two of the six are smaller siblings of
others from the same provider, all three providers train on overlapping web
corpora, and every variant received identical prompt text, so agreement across
them is weaker evidence than twelve independent elicitations would be.

\begin{table}[htbp]
\centering
\footnotesize
\caption{The twelve elicitation variants. ``Mode'' distinguishes a plain
completion from one in which the model was asked to reason before emitting the
coefficient block; the two modes use different JSON schemas, described in
Appendix~\ref{secA:prompt}. ``Stored as'' is the directory under
\texttt{priors/<dataset>/} in the repository. Dated API snapshot identifiers
were not recorded at elicitation time; see the note below.}
\label{tab:variants}
\begin{tabular}{@{}lllll@{}}
\toprule
Label & Provider & Model & Mode & Stored as \\
\midrule
GPT-5.1          & OpenAI    & GPT-5.1          & plain     & \texttt{openai\_\_gpt-5-1} \\
GPT-5.1 (R)     & OpenAI    & GPT-5.1          & reasoning & \texttt{openai\_\_gpt-5-1\_\_reasoning} \\
GPT-5-mini       & OpenAI    & GPT-5-mini       & plain     & \texttt{openai\_\_gpt-5-mini} \\
GPT-5-mini (R)  & OpenAI    & GPT-5-mini       & reasoning & \texttt{openai\_\_gpt-5-mini\_\_reasoning} \\
Opus 4.5         & Anthropic & Claude Opus 4.5  & plain     & \texttt{anthropic\_\_opus-4-5} \\
Opus 4.5 (R)    & Anthropic & Claude Opus 4.5  & reasoning & \texttt{anthropic\_\_opus-4-5\_\_reasoning} \\
Haiku 4.5        & Anthropic & Claude Haiku 4.5 & plain     & \texttt{anthropic\_\_haiku-4-5} \\
Haiku 4.5 (R)   & Anthropic & Claude Haiku 4.5 & reasoning & \texttt{anthropic\_\_haiku-4-5\_\_reasoning} \\
Gemini Pro       & Google    & Gemini 2.5 Pro   & plain     & \texttt{google\_\_gemini-2-5-pro} \\
Gemini Pro (R)  & Google    & Gemini 2.5 Pro   & reasoning & \texttt{google\_\_gemini-2-5-pro\_\_reasoning} \\
Gemini Flash     & Google    & Gemini 2.5 Flash & plain     & \texttt{google\_\_gemini-2-5-flash} \\
Gemini Flash (R) & Google    & Gemini 2.5 Flash & reasoning & \texttt{google\_\_gemini-2-5-flash\_\_reasoning} \\
\bottomrule
\end{tabular}
\end{table}

One reproducibility caveat belongs with this table. The elicitation recorded
the model family but not the dated snapshot identifier the provider served at
the time, so the exact weights behind each bundle cannot be determined and
re-querying the same family later will not reproduce the stored draws. This is
the reason the samples themselves are archived rather than a script for
regenerating them: the $60$ stored bundles, not the model names, are the
reproducible artefact. Future work of this kind should log the full versioned
identifier returned by each API call.

For each (dataset, variant) combination, $M = 50$ samples were drawn at temperature $0.8$: high enough for meaningful diversity, low enough to avoid incoherence. Outputs with missing keys, extra keys or invalid numeric values were rejected and replaced by additional calls until $50$ valid samples accumulated, subject to a cap on total attempts. A single prompt variant is used for every LLM and dataset, to avoid overfitting the elicitation pipeline to these datasets and splits. Appendix~\ref{secA:prompt} reproduces a complete system and user prompt verbatim.

\subsection{Prior quantities and model fitting}

The elicited vectors are the $\beta^{(m)}$ of Eq.~\ref{eq:priorquant}, treated as draws from the model's implicit belief distribution over the coefficients, and $\widehat{\mu}$, $\widehat{\sigma}^{2}_{j}$ and $\widehat{\Sigma}$ are computed from them exactly as defined there. The precision $\widehat{P} = \widehat{\Sigma}^{-1}$ is formed with the diagonal jitter described there.

Regression solutions to Eq.~\ref{eq:map} are closed-form; logistic solutions use L-BFGS, with Correlated-Prior started at zero and the other families at a seeded $0.01\,\mathcal{N}(0, 1)$ draw. We note that the Correlated-Prior logistic fits are sensitive to the optimiser path: $\widehat{\Sigma}$ estimated from $M = 50$ draws is ill-conditioned, and its inverse yields a penalty under which L-BFGS terminates at slightly different points depending on the starting value and on whether the fit is warm-started from the $\lambda$ grid walk. The resulting differences are in the third decimal of the reported metrics and do not affect any conclusion, but they are a further argument for shrinking $\widehat{\Sigma}$ before inversion.

\subsection{Tuning, selection and evaluation}

The $\lambda$ grid is given as $51$ log-spaced values from $10^{-4}$ to $10^{6}$; the $\gamma$ grid is $101$ equally spaced values on $[0,1]$. For classification, the baseline is tuned on $\lambda_{\text{equiv}} = 1/C$ such that its regularisation axis runs in the same direction as the one of prior families. Hyperparameters are chosen by 5-fold cross-validation on the source training pool under a strict-best-mean rule: for each grid value, per-fold validation scores are averaged and the best average selected (lowest MSE for regression, highest AUC for classification). Folds are built once per dataset and reused by every variant, so all variants are scored on identical splits. For each family, the variant whose family model achieves the best mean CV score is deployed; this within-family rule produces the four models reported per dataset in Tables~\ref{tab:regression} and~\ref{tab:classification}. Where a single model is required, as in the selection analysis of Section~\ref{subsec:selection}, the same criterion is applied across the whole grid of $48$ candidates, selecting the one (variant, family) combination with the best mean CV score. Both rules use source data only. We also record the fold-level standard error of the CV score at the selected candidate, because the margins between competing candidates turn out to be small relative to it. Models are then refit on the full training pool and evaluated on the held-out ID test and OOD sets. The OOD set never enters any tuning or selection step.

Regression reports MSE. Classification reports AUC, log loss and Brier score as threshold-free metrics, and accuracy, precision, sensitivity, specificity and $F_1$ at a threshold.

One bootstrap procedure is used in two places and it is worth separating them. Within a single fitted deployment, uncertainty is assessed by $2{,}000$ resamples of the OOD test set, stratified by class for classification, with percentile intervals from the $2.5$th and $97.5$th percentiles. The same resample indices are applied to every model within a call, so per-resample draws are paired; improvement-over-baseline claims use the paired difference $\dd = \text{model} - \text{baseline}$, which accounts for the correlation between a prior-informed model and its baseline on the same OOD sample. Taken alone this quantifies sensitivity to which OOD rows happened to be observed and, by construction, not variability in training, tuning or elicitation. It is \emph{not} the interval reported in the main results: those nest this resampling inside repeated source splits, as described under ``Uncertainty intervals'' below. The single-deployment version is what Appendix~\ref{secA:threshold} uses for the threshold analysis, where the quantity of interest is a comparison at one fitted operating point.

\subsection{Follow-up analyses}

\paragraph{Threshold recalibration.}
A fixed $0.5$ cut-off is not a neutral basis for comparison when the domain split moves outcome prevalence, because it rewards whichever model happens to place more probability mass above it rather than the model that ranks or calibrates better. To compare models fairly we therefore give every one of them, the baseline included, the same freedom to choose an operating point, and we choose it on source data only. For each classification dataset and each deployed model including the baseline, a threshold is selected from $91$ values on $[0.05, 0.95]$ by maximising mean 5-fold CV $F_1$ on the source training pool. Models are then re-scored on the OOD set at their own selected threshold. Paired bootstrap intervals on $\dd F_1$ are computed with each model at its own threshold. The OOD-maximising threshold is also recorded as a diagnostic upper bound.

\paragraph{Placebo priors.}
The placebos test whether the elicited numbers contain usable information about \emph{these particular} features, or whether any shrinkage target of roughly the same shape and magnitude would serve as well. Each one corrupts the prior's content while preserving its structure. Permuted placebos draw a random permutation $\pi$ of the $p$ features and re-index the whole prior: $\muhat \mapsto \muhat_\pi$, $\widehat{\sigma}^2 \mapsto \widehat{\sigma}^2_\pi$, $\Phat \mapsto \Phat_{\pi\pi}$. Sign-flip placebos draw $s \in \{-1, +1\}^p$ uniformly and set $\muhat \mapsto \muhat \odot s$, leaving variances unchanged and mapping $\Phat \mapsto D\Phat D$ with $D = \mathrm{diag}(s)$, which is the correct transformation since $D^{-1} = D$ and $\Sigma \mapsto D \Sigma D$. Both preserve the prior's scale and its correlation magnitudes. For every draw, $\lambda$ is re-tuned by the same CV rule, so placebos receive the same tuning advantage as the real prior. We use $200$ placebo draws per (dataset, family, placebo type) combination for regression and $100$ for classification, and report the empirical $p$-value $(\#\{\text{placebo} \ge \text{real}\} + 1)/(n+1)$. The Adult dataset is excluded: cross-validation drives $\lambda$ to the grid floor there and the prior families collapse onto the baseline, so a placebo comparison is meaningless by construction.

\paragraph{Uncertainty intervals.}
Every interval in the Results propagates three sources of variability together.
The ID $80/20$ split is redrawn $S$ times ($S = 50$ for Boston Housing and
Cancer, $30$ for Heart Disease and Diabetes, $10$ for Adult); inside each
redraw $\lambda$ is re-selected by the cross-validation rule above and the
model refitted; and the resulting model is then scored on $200$ bootstrap
resamples of the OOD partition, stratified by class for classification. The
$S \times 200$ resulting values of the lift over the baseline are pooled.

Two intervals are read off this construction and they answer different
questions. The $2.5$th to $97.5$th percentile range of the pooled draws is a
\emph{prediction interval}: it describes the lift a single future run of the
whole pipeline would realise, and the accompanying fraction is the share of
those draws below zero. A \emph{confidence interval for the mean lift} is
narrower, because the source-split component is averaged over the $S$
repetitions while the target component is not, all $S$ splits are scored on
the same realised OOD partition, so there is only one draw of the target
sample. Writing $\sigma^{2}_{\text{split}}$ for the variance of the $S$
per-split means and $\sigma^{2}_{\text{tgt}}$ for the mean within-split
bootstrap variance, we report
\begin{equation}
\mathrm{SE}(\bar{\dd}) \;=\; \sqrt{\sigma^{2}_{\text{split}}/S \;+\;
\sigma^{2}_{\text{tgt}}},
\label{eq:se}
\end{equation}
and a normal interval $\bar{\dd} \pm 1.96\,\mathrm{SE}(\bar{\dd})$. The
distinction matters: read as a confidence interval, a prediction interval
understates the evidence substantially, because $\sigma^{2}_{\text{split}}$
enters at full width rather than divided by $S$. Resampling
only the OOD partition would capture target-sample variability alone, and
redrawing only the ID split would capture training and tuning variability
alone; the two are disjoint, so both are needed. Elicitation variability is not
covered, as the deployed variant and its $M = 50$ draws are held fixed.

\paragraph{Resplits.}
The ID $80/20$ split is redrawn under $50$ seeds (Boston Housing, Cancer), $30$ (Heart Disease, Diabetes) or $10$ (Adult), and the full deployment step is repeated with $\lambda$ re-tuned from scratch each time. The OOD partition is fixed by the domain split, so only the training sample varies. The selected variant per family is held at the deployed choice, so this measures stability of the lift given that variant rather than stability of variant selection. Adult uses every fourth $\lambda$ grid point for tractability, since each logistic fit on $n \approx 27{,}000$ costs about two seconds and its $\lambda$ is driven to the lowest grid value regardless.



\section{Results}

We report predictive performance first on the two regression datasets and then on the three classification datasets, followed by a series of diagnostic controls. These controls are necessary because raw predictive metrics alone cannot explain why a model improves, nor whether the approach is dependable in practice. An observed lift could simply be an artifact of shrinking toward an arbitrary non-zero target rather than genuine domain understanding, while reporting only the final selected model conceals both the risk of harmful priors and the noise inherent in source-domain model selection. We therefore test whether the benefit remains under systematic corruption of the priors, assess how much provider and prior-structure choices matter, and examine what the elicited priors actually agree on. The regression subsection also sets out how every interval in the paper is computed, and the later subsections refer back to it.

\subsection{OOD prediction on regression tasks}

Table~\ref{tab:regression} reports the deployed model per family on the two regression datasets, with two intervals that answer different questions. The \emph{confidence interval} is for the mean lift: if the whole procedure were repeated many times, how precisely do we know the average benefit? The \emph{prediction interval} is for one run: if a practitioner deploys this pipeline once, how much could the realised lift vary? The two differ because the source-split component of variance is averaged over the $S$ repetitions in the first and enters at full width in the second, and the distinction matters here: the prediction interval covers zero almost everywhere, the confidence interval does not.

On Boston Housing the Independent-Prior and Correlated-Prior are the strongest: Correlated-Prior with Opus~4.5 averages a $7.9\%$ reduction in OOD MSE and Independent-Prior $7.4\%$, while Prior-Mean averages essentially nothing ($-0.2\%$). All three shrink toward the same elicited mean $\muhat$ and differ only in the prior covariance, so what separates them is the treatment of spread. Prior-Mean assigns every coefficient the common prior variance $\lambda^{-1}$ and discards the per-feature spread entirely; Independent-Prior takes each coefficient's variance from the observed spread of the $M$ draws; Correlated-Prior additionally uses the cross-coefficient covariance. On Boston Housing it is therefore not the elicited location that produces the effect (Prior-Mean has that and gains nothing from it), but the elicited \emph{relative confidence}.

In fact, this elicited confidence is far more rigid than a standard variance metric suggests: rather than exhibiting a narrow spread, the LLM returned bit-identical values across all fifty draws for four of the eleven features. Consequently, the elicited standard deviation is exactly zero and the penalty weight $\lambda/(\widehat{\sigma}^{2}_{j} + \varepsilon)$ is set by the floor $\varepsilon$ rather than by the elicitation. Those four coefficients are effectively held at $\muhat_{j}$. This is a strong modelling commitment and we test it rather than assert it: Table~\ref{tab:epsilon} re-tunes and refits across four orders of magnitude of $\varepsilon$ and against a Ledoit--Wolf shrinkage estimator of $\Sighat$ in place of the jitter. The Boston Housing lift is $+11.15\%$ at the deployed setting and $+11.10\%$ under Ledoit--Wolf, and degrades only as $\varepsilon$ grows large enough to erase the elicited spread altogether. The effect is a property of the elicitation, not of the floor.

The reverse holds on Cancer, where Prior-Mean averages $+2.4\%$ and Independent-Prior and Correlated-Prior around $+1\%$. This is not a case of the elicited spread being uninformative for lack of variation: the deployed variants there have no zero-variance coefficients at all, and their standard deviations range over a factor of two to four around the median. It is worth noting when the families would become identical: if the elicited standard deviations were equal across features, $\mathrm{diag}(1/\widehat{\sigma}^{2}_{j})$ would reduce to a multiple of the identity and Independent-Prior would be Prior-Mean with $\lambda$ rescaled, which the tuning grid absorbs. They separate only to the extent that the elicitation expresses differential confidence, and on Cancer the differential confidence it expressed was actively unhelpful, the coefficients the models were most certain about were not the ones that transferred best.

Taken together the two datasets separate two things the elicitation supplies. The location $\muhat$ is useful on both. The relative confidence attached to it is useful on Boston Housing and misleading on Cancer, which is why no single prior structure is best on both datasets and why the choice between them has to be made per dataset rather than settled once.

Five of the six prior-informed results have positive means, and under the confidence interval five of six exclude zero: Independent-Prior and Correlated-Prior on Boston Housing and all three families on Cancer. Only Boston Housing's Prior-Mean, which gains nothing on that dataset, does not.

\begin{table}[htbp]
\centering
\footnotesize
\setlength{\tabcolsep}{2pt}
\caption{Regression datasets. The selected variant and $\lambda$ are those
chosen by mean 5-fold CV on the source training pool of the reference split, and
the OOD MSE column is that deployed model's error on the OOD partition. The MSE
and the lift columns come from two different computations: the MSE is a
single number from the reference split, whereas the lift and both intervals are
averaged over repeated source splits, so the lift is not
recoverable by comparing the two MSE entries. Lift is the percentage
reduction in OOD MSE relative to the ridge baseline. ``CI'' is a $95\%$
confidence interval for the \emph{mean} lift; ``PI'' is a $95\%$ prediction
interval for the lift of a \emph{single} deployment. Both propagate
source-split, $\lambda$-selection and target-sample variability; they differ in
whether the source component is averaged over the $S$ splits. ``Below'' is the
fraction of prediction draws in which the prior-informed model is worse than
baseline. The variant named is the CV-best one the pipeline deploys, not the
OOD-best one. A dagger ($^{\dagger}$) marks a confidence interval excluding
zero.}
\label{tab:regression}
\begin{tabular}{@{}llrrrllr@{}}
\toprule
Family & Selected variant & $\lambda$ & OOD MSE & Lift & 95\% CI (mean) & 95\% PI (one run) & Below \\
\midrule
\multicolumn{8}{@{}l}{\emph{Boston Housing} ($n_{\text{train}}=284$, $n_{\text{OOD}}=151$)}\\
Unregularised ref. & --             & --    & 44.99 & --      & --   & --  & --  \\
Baseline ridge     & --             & 25.1  & 44.10 & --      & --   & --  & --  \\
Prior-Mean         & Gemini Pro     & 251   & 42.92 & $-0.18$ & [$-8.29$, $+7.92$] & [$-12.97$, $+10.54$] & 48\% \\
Independent-Prior  & Opus 4.5       & 10.0  & 39.56 & $+7.40$ & [$+0.57$, $+14.24$]$^{\dagger}$ & [$-1.04$, $+18.23$]  & 4\% \\
Correlated-Prior   & Opus 4.5       & 15.8  & 39.18 & $+7.85$ & [$+1.18$, $+14.51$]$^{\dagger}$ & [$-0.61$, $+18.87$]  & 3\% \\
Mixed              & Gemini Pro     & $\gamma{=}1.00$ & 42.92 & $-0.18$ & [$-8.29$, $+7.92$] & [$-12.97$, $+10.54$] & 48\% \\
\midrule
\multicolumn{8}{@{}l}{\emph{Cancer} ($n_{\text{train}}=1{,}106$, $n_{\text{OOD}}=1{,}664$)}\\
Unregularised ref. & --             & --    & 524.17 & --      & --   & --  & --  \\
Baseline ridge     & --             & 100   & 501.22 & --      & --   & --  & --  \\
Prior-Mean         & Opus 4.5       & 158   & 481.67 & $+2.37$ & [$+1.71$, $+3.04$]$^{\dagger}$ & [$-0.38$, $+7.53$] & 12\% \\
Independent-Prior  & GPT-5.1        & 158   & 487.27 & $+1.13$ & [$+0.60$, $+1.66$]$^{\dagger}$ & [$-1.24$, $+3.44$] & 11\% \\
Correlated-Prior   & Gemini Pro (R) & 158   & 490.60 & $+0.90$ & [$+0.43$, $+1.38$]$^{\dagger}$ & [$-1.11$, $+3.38$] & 17\% \\
Mixed              & Opus 4.5       & $\gamma{=}0.68$ & 487.69 & $+1.65$ & [$+1.19$, $+2.11$]$^{\dagger}$ & [$-0.26$, $+5.30$] & 11\% \\
\bottomrule
\end{tabular}
\end{table}

The same models were also scored on the held-out in-domain test split, and the comparison bears directly on what the prior is doing. On Boston Housing the in-domain gain is \emph{larger} than the out-of-domain gain: $+13.2\%$ against $+11.2\%$ for Correlated-Prior on the reference split. The same holds on Heart Disease ($+1.7\%$ in-domain against $+1.0\%$) and Diabetes ($+3.7\%$ against $+1.0\%$). On those three datasets the elicited prior is a better regularisation rather than a correction for domain shift specifically. Cancer is the exception: in-domain the prior is worth $+0.1\%$ and out-of-domain $+3.9\%$, so there the benefit is specific to the shift. Section~\ref{subsec:whenhelps} returns to why Cancer differs.

The construction comes with two caveats. First, bootstrapping the OOD partition treats it as a sample from $P_{T}$, which is the right way to view the risk $R_{T}$ of Eq.~\ref{eq:risk}, but each OOD partition here is a \emph{complete} subpopulation: every non-Southern county, every Hungarian and Swiss patient in the database, rather than a draw from a larger pool. For the narrower question ``did the prior help on this particular deployment population'', the target is fully observed and the $\sigma^{2}_{\text{tgt}}$ term of Eq.~\ref{eq:se} should be dropped. Doing so narrows every interval and adds six further results that exclude zero, among them all four Diabetes log-loss and Prior-Mean results. We report the generalising version throughout because it is the question a practitioner faces before deploying; Appendix~\ref{secA1} shows how the estimated performance gain varies across source-data splits. For each split, we refit the models but evaluate them on the same OOD sample. Averaging over splits therefore reduces uncertainty due to the source split, but it does not provide additional target observations or reduce uncertainty due to the finite OOD sample. This distinction matters most for Boston Housing: with only 151 OOD observations, target-sample uncertainty accounts for roughly half of the variance used to construct its confidence intervals. Cancer has 1,664 OOD observations, so this component accounts for less than 5\% of its interval variance. Its confidence intervals are consequently much narrower, even though its estimated gain is smaller.

\subsection{OOD prediction on classification tasks}

Table~\ref{tab:classification} shows the predictive performance of the different models on the three classification datasets. Two metrics are reported for each: AUC, which measures only how well the model ranks observations, and log loss, which measures the quality of the predicted probabilities themselves. Separating the two turns out to be important, because the priors act on one and not the other.

Ranking is barely affected on Heart Disease, where all three AUC confidence intervals include zero. Proper scoring rules tell a different story: log loss improves by more than 10\% for every family on Heart Disease ($0.4709 \rightarrow 0.4327$ for Prior-Mean), and all three confidence intervals exclude zero (+12.23\% for Prior-Mean and +12.43\% for Correlated-Prior). Diabetes improves on both metrics, with AUC confidence intervals excluding zero for Independent-Prior and Correlated-Prior, alongside positive log-loss gains of 3–4\%. Brier scores move consistently in the same direction, and reliability curves show that the prior-informed models track calibration more closely than the baseline (Supplementary Fig.~\ref{fig:calibration}). Adult, by contrast, is the exception to these gains: with $n_{\text{train}} \approx 27{,}000$, the baseline model is already close to its best attainable accuracy, cross-validation drives $\lambda$ to the grid floor, and the prior families collapse onto the baseline. While two Adult results technically exclude zero in the direction of harm ($-0.01\%$ AUC and $-0.05\%$ log loss), they are detectable only because $n_{\text{OOD}} = 14{,}936$ renders them statistically precise, and have no practical consequence.

Threshold-dependent metrics such as $F_1$, sensitivity, and specificity are deliberately omitted from Table ~\ref{tab:classification}. All such metrics require an operating cut-off, but the domain splits here induce substantial label shift: outcome prevalence shifts from 0.212 to 0.495 on Diabetes and from 0.459 to 0.530 on Heart Disease. Under such shifts, evaluating models at a fixed 0.5 threshold artificially rewards whichever model produces higher probabilities, regardless of calibration or ranking quality. Appendix~\ref{secA:threshold} demonstrates how this prevalence shift manufactures an apparent $F_1$ gain that disappears once baselines are granted the same threshold freedom, confirming that proper scoring rules provide the only defensible basis for classification evaluation here.

\begin{table}[htbp]
\centering
\footnotesize
\setlength{\tabcolsep}{2pt}
\caption{Classification datasets. ``Value'' is the deployed model's metric on
the reference split; ``lift'' is the percentage improvement over the
$L_{2}$-logistic baseline. ``CI'' is a $95\%$ confidence interval for the mean
lift and ``PI'' a $95\%$ prediction interval for a single deployment; both
propagate source-split, $\lambda$-selection and target-sample variability and
differ only in whether the source component is averaged over the $S$ splits.
``Below'' is the fraction of prediction draws in which the prior-informed model
is worse than baseline. The Mixed family is omitted and tracks Prior-Mean
closely ($\gamma = 1.00$ and $0.97$). A dagger ($^{\dagger}$) marks a
confidence interval excluding zero; on Adult the two such entries point in the
direction of harm, at magnitudes of no practical consequence.}
\label{tab:classification}
\begin{tabular}{@{}lllrrllr@{}}
\toprule
Family & Variant & Metric & Value & Lift & 95\% CI (mean) & 95\% PI (one run) & Below \\
\midrule
\multicolumn{8}{@{}l}{\emph{Heart Disease} ($n_{\text{train}}=242$, $n_{\text{OOD}}=417$)}\\
Baseline          & --              & AUC     & 0.8779 & --       & --  & --  & --  \\
Prior-Mean& Opus 4.5        & AUC     & 0.8860 & $+0.08$  & [$-0.75$, $+0.91$] & [$-1.13$, $+2.15$] & 51\% \\
Independent-Prior& Gemini Pro (R) & AUC     & 0.8845 & $-0.07$  & [$-1.13$, $+1.00$] & [$-1.50$, $+1.90$] & 57\% \\
Correlated-Prior& Opus 4.5        & AUC     & 0.8862 & $+0.18$  & [$-0.68$, $+1.04$] & [$-1.03$, $+2.47$] & 47\% \\
Baseline          & --              & LogLoss & 0.4709 & --       & --  & --  & --  \\
Prior-Mean& Opus 4.5        & LogLoss & 0.4327 & $+12.23$ & [$+4.20$, $+20.25$]$^{\dagger}$ & [$-1.68$, $+39.30$] & 8\% \\
Independent-Prior& Gemini Pro (R) & LogLoss & 0.4407 & $+10.78$ & [$+3.20$, $+18.36$]$^{\dagger}$ & [$-3.45$, $+37.97$] & 19\% \\
Correlated-Prior& Opus 4.5        & LogLoss & 0.4331 & $+12.43$ & [$+4.85$, $+20.01$]$^{\dagger}$ & [$-1.45$, $+39.43$] & 7\% \\
\midrule
\multicolumn{8}{@{}l}{\emph{Diabetes} ($n_{\text{train}}=316$, $n_{\text{OOD}}=372$)}\\
Baseline          & --              & AUC     & 0.7730 & --       & --  & --  & --  \\
Prior-Mean& Opus 4.5        & AUC     & 0.7796 & $+1.41$  & [$-0.11$, $+2.93$] & [$-0.28$, $+3.54$] & 5\% \\
Independent-Prior& Opus 4.5        & AUC     & 0.7815 & $+1.78$  & [$+0.12$, $+3.44$]$^{\dagger}$ & [$-0.24$, $+4.10$] & 4\% \\
Correlated-Prior& Opus 4.5        & AUC     & 0.7805 & $+1.81$  & [$+0.16$, $+3.47$]$^{\dagger}$ & [$-0.24$, $+4.14$] & 4\% \\
Baseline          & --              & LogLoss & 0.6327 & --       & --  & --  & --  \\
Prior-Mean& Opus 4.5        & LogLoss & 0.6025 & $+3.44$  & [$-0.38$, $+7.25$] & [$-1.57$, $+23.44$] & 17\% \\
Independent-Prior& Opus 4.5        & LogLoss & 0.6047 & $+3.50$  & [$-0.52$, $+7.51$] & [$-1.79$, $+23.39$] & 15\% \\
Correlated-Prior& Opus 4.5        & LogLoss & 0.6035 & $+3.28$  & [$-0.74$, $+7.30$] & [$-2.01$, $+23.39$] & 18\% \\
\midrule
\multicolumn{8}{@{}l}{\emph{Adult} ($n_{\text{train}}=27{,}124$, $n_{\text{OOD}}=14{,}936$)}\\
Baseline          & --              & AUC     & 0.8770 & --       & --  & --  & --  \\
Prior-Mean& GPT-5-mini      & AUC     & 0.8771 & $-0.00$  & [$-0.00$, $+0.00$] & [$-0.01$, $+0.00$] & 64\% \\
Independent-Prior& Gemini Flash (R) & AUC     & 0.8768 & $-0.01$  & [$-0.02$, $-0.01$]$^{\dagger}$ & [$-0.04$, $-0.01$]$^{\dagger}$ & 100\% \\
Correlated-Prior& Opus 4.5 (R)   & AUC     & 0.8769 & $-0.01$  & [$-0.03$, $+0.00$] & [$-0.07$, $+0.00$] & 96\% \\
Baseline          & --              & LogLoss & 0.3940 & --       & --  & --  & --  \\
Prior-Mean& GPT-5-mini      & LogLoss & 0.3940 & $+0.00$  & [$-0.01$, $+0.02$] & [$-0.02$, $+0.06$] & 56\% \\
Independent-Prior& Gemini Flash (R) & LogLoss & 0.3942 & $-0.05$  & [$-0.08$, $-0.02$]$^{\dagger}$ & [$-0.10$, $-0.02$]$^{\dagger}$ & 100\% \\
Correlated-Prior& Opus 4.5 (R)   & LogLoss & 0.3941 & $-0.04$  & [$-0.09$, $+0.01$] & [$-0.19$, $+0.01$] & 91\% \\
\bottomrule
\end{tabular}
\end{table}

Adult deserves a more careful reading than ``the prior does not help when data are plentiful''. There are two distinct reasons a prior might be silent at large $n$, and they have different implications. The first is statistical: a prior of fixed precision contributes a constant to the log-posterior while the likelihood contributes a term growing with $n$, so its influence decays as $1/n$ and the posterior concentrates on the maximum-likelihood estimate. Under domain shift that estimate converges to the source optimum $\beta^{\star}_{S}$, whose bias with respect to the target does \emph{not} shrink with $n$; a large source sample therefore gives an increasingly precise estimate of the wrong quantity while at the same time removing the one source of information that could correct it. On that argument a prior ought to remain useful on Adult, since the bias it would correct is a property of the domain rather than of the sample size.

The second reason is specific to how we set the prior weight. Because $\lambda$ is tuned rather than fixed, the asymptotic argument does not bind automatically, $\lambda$ could in principle grow with $n$, but it is selected to minimise source-domain cross-validation error, and a prior that trades in-domain fit for out-of-domain transfer is strictly bad by that criterion. On Adult, the cross-validation drives the prior weight low enough that every family collapses onto the baseline, and at that point the two reasons are observationally equivalent: a prior that contains nothing useful for this dataset and a prior whose usefulness the selection rule cannot see produce the same null. Separating them would require a selection objective aimed at the target domain rather than the source, which is a change to the method rather than a further analysis of it, and we take it up in the Discussion. That the prior does no harm here is still the property the design was meant to guarantee: two of Adult's intervals formally exclude zero on the side of harm, but at $-0.01\%$ AUC and $-0.05\%$ log loss, detectable only because $14{,}936$ OOD rows make them precise.

\subsection{Placebo priors: the priors' content, not their scale}

A prior-informed model differs from ridge in two ways at once: it shrinks toward a non-zero target, and that target makes specific claims about which features matter and in which direction. The permuted and sign-flip placebos defined in the Methods separate these two properties: each preserves the prior's overall scale and its correlation magnitudes while attaching them to the wrong features or the wrong directions. A model that gains merely from shrinking toward something non-zero of about the right size should therefore do as well under a placebo as under the real prior.

\begin{figure}[htbp]
\centering
\includegraphics[width=\linewidth]{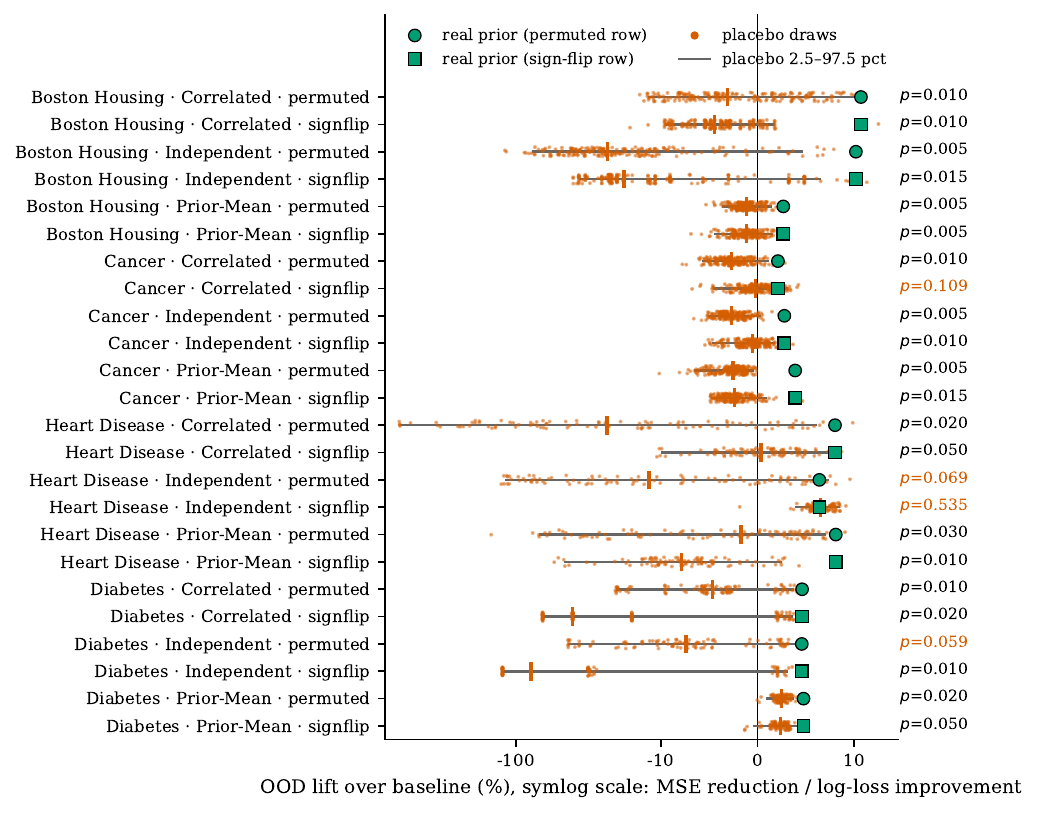}
\caption{Placebo priors. Each row is one (dataset, prior family, placebo type)
combination. Vermillion points are individual placebo draws ($200$ for regression,
$100$ for classification) with the grey bar marking their $2.5$--$97.5$
percentile range and the tick their median; the bluish-green marker is the real
elicited prior. $p$ is the empirical fraction of placebo draws reaching at
least the real prior's lift. Lift is OOD MSE reduction for Boston Housing and
Cancer, and OOD log-loss improvement for Heart Disease and Diabetes, the
metrics the priors actually move, as established in
Tables~\ref{tab:regression} and~\ref{tab:classification}. The horizontal axis
is symmetric-log beyond $\pm 10\%$, because corrupted priors on the
classification datasets can be worse than baseline by several hundred per
cent.}
\label{fig:placebo}
\end{figure}

Figure~\ref{fig:placebo} reports the outcome for all $24$ (dataset, prior family, placebo) combinations. The real prior beats its placebo distribution in $20$ of them at $p < 0.05$, and the two corruptions are equally damaging: permutation, which keeps every elicited number but attaches it to the wrong feature, is rejected in $10$ of $12$ comparisons, and sign-flipping, which keeps each magnitude and discards only the direction, in $10$ of $12$ as well. The effect therefore depends on both the identity of the features the prior refers to and the directions it assigns to them.

The four exceptions do not group by placebo type but by family: three of them are Independent-Prior comparisons (Heart Disease under both corruptions, $p = 0.069$ and $p = 0.535$, and Diabetes under permutation, $p = 0.059$), with Correlated-Prior sign-flip on Cancer ($p = 0.109$) the fourth. By dataset the counts are six of six on Boston Housing, five of six on Cancer and Diabetes, and four of six on Heart Disease. On the regression datasets every placebo median is \emph{below} the ridge baseline, and on Boston Housing at most two of $200$ draws match the real lift in any comparison, so a corrupted prior there is not merely less useful than the real one but actively worse than no prior at all.

The choice of metric matters here, and it is worth being explicit about it. The classification rows are scored on log loss rather than AUC, for the same reason Table~\ref{tab:classification} leads with log loss: AUC is the metric the priors leave essentially flat, so testing the priors' content against it asks whether a corruption damages something the elicitation was never moving. It does not, and the result is correspondingly weak: on AUC only $7$ of $12$ classification comparisons reject, and sign-flipping is not rejected in four of six. That contrast is itself a caution for this literature. A placebo or ablation evaluated on a threshold-free ranking metric can understate an effect that is real in the predicted probabilities, and the mean real lift on these datasets is $+0.91\%$ on AUC against $+6.08\%$ on log loss.

The placebo tests clarify why the elicited priors help. Permuting the prior values across features or changing their signs generally reduces their predictive benefit, even though the values’ overall scale is preserved. The gains therefore depend on assigning plausible effects to the appropriate features and in the appropriate directions—not simply on shrinking coefficients toward non-zero values.

\subsection{What the selection step can and cannot do}\label{subsec:selection}

\begin{figure}[htbp]
\centering
\includegraphics[width=0.9\linewidth]{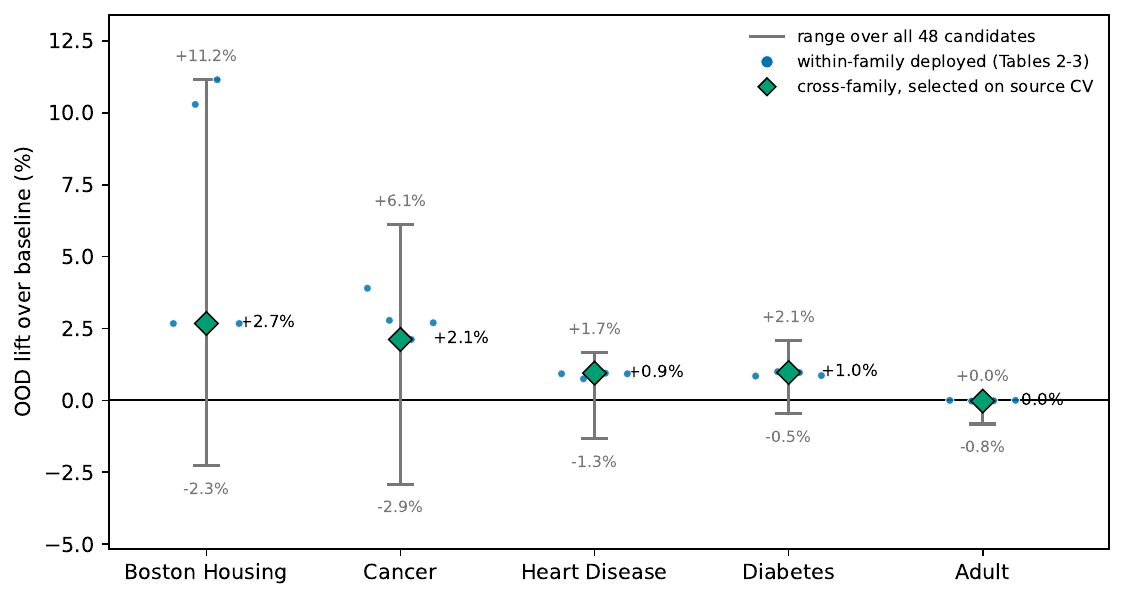}
\caption{Per-dataset OOD lift over baseline across all $48$ (12 elicitation variants $\times$ 4 prior model families)
candidates at their CV-tuned $\lambda$. Grey bar: the range from worst to best
candidate. Blue circles: the four models deployed by the within-family rule,
which are the models reported in Tables~\ref{tab:regression}
and~\ref{tab:classification}. Green diamond: the single model selected by a
cross-family rule, the best of all $48$ candidates by source-domain
cross-validation. Every quantity is computed from source data alone; the
distance from each marker to the top of its bar is the benefit that
source-domain selection does not reach.}
\label{fig:selection}
\end{figure}

Every result above reports on a selected model, thus the properties of the selection rule matter as much as the properties of the priors. Per dataset there are $48$ candidate combinations (12 language model $\times$ 4 prior model families), each at its own CV-tuned $\lambda$. Figure~\ref{fig:selection} shows the full range of OOD lifts over that combinations, the four models the within-family rule deploys, and the single model a cross-family rule selects.

The spread is wide: the candidate pool ranges from $-2.3\%$ to $+11.2\%$ on Boston Housing and from $-2.9\%$ to $+6.1\%$ on Cancer, and the worst candidate is below baseline on every dataset. An LLM-elicited prior is therefore not automatically safe because it is domain-informed. A study using a single provider, prompt and prior structure could have concluded that the method is clearly beneficial, neutral or actively harmful depending only on which combination it happened to pick, and that observation is the strongest practical argument in this paper for reporting the whole candidate distribution rather than a deployed configuration.

Source-domain selection uses that spread only partially. A cross-family rule picks Gemini~Pro with Prior-Mean on Boston Housing, worth $+2.7\%$ compared with the best achievable $+11.2\%$, and Gemini~Pro~(R) with Correlated-Prior on Cancer, worth $+2.1\%$ against $+6.1\%$. On the classification datasets it reaches $+0.9\%$ of $+1.7\%$ and $+1.0\%$ of $+2.1\%$. In every case the selected candidate is in the middle of the candidate range rather than near the best.

Source-domain cross-validation does not clearly distinguish the leading candidates. We assess the gap between the winner and the next distinct candidate using the standard error of their \emph{paired} score difference. All candidates are evaluated on the same five folds, so a difficult fold tends to affect both scores, and much of that variation cancels when they are compared. The gap is only $0.06$ standard errors on Boston Housing, Cancer and Adult, and $0.25$ on Heart Disease; it reaches $1.63$ on Diabetes. Thus, except perhaps on Diabetes, the cross-validation scores provide little basis for treating the selected candidate as clearly better than its closest alternative.

On Boston Housing, the two highest-scoring entries are actually the same fitted model: the Mixed family selects $\gamma = 1$, making it equivalent to Prior-Mean. We therefore compare the winner with the next \emph{distinct} candidate. Although selection does not reliably identify the best-performing prior, the selected models avoid the worst-performing candidates in these experiments. Tables~\ref{tab:regression} and~\ref{tab:classification} report the variants selected within each prior family, as prescribed by the deployment procedure; their reported gains should not be interpreted as the result of reliably finding the best candidate.

What the selection step contributes is therefore protection rather than optimisation. Together with the tuned $\lambda$ it keeps the deployed model away from the worst-performing candidates on every dataset, and on Adult, where the prior contains nothing useful, cross-validation drives the weight low enough that every family collapses onto the baseline and the cost of a useless prior is $-0.03\%$. That is a safety property, not an optimality property.

\subsection{What the elicited priors agree on}

\begin{figure}[htbp]
\centering
\includegraphics[width=\linewidth]{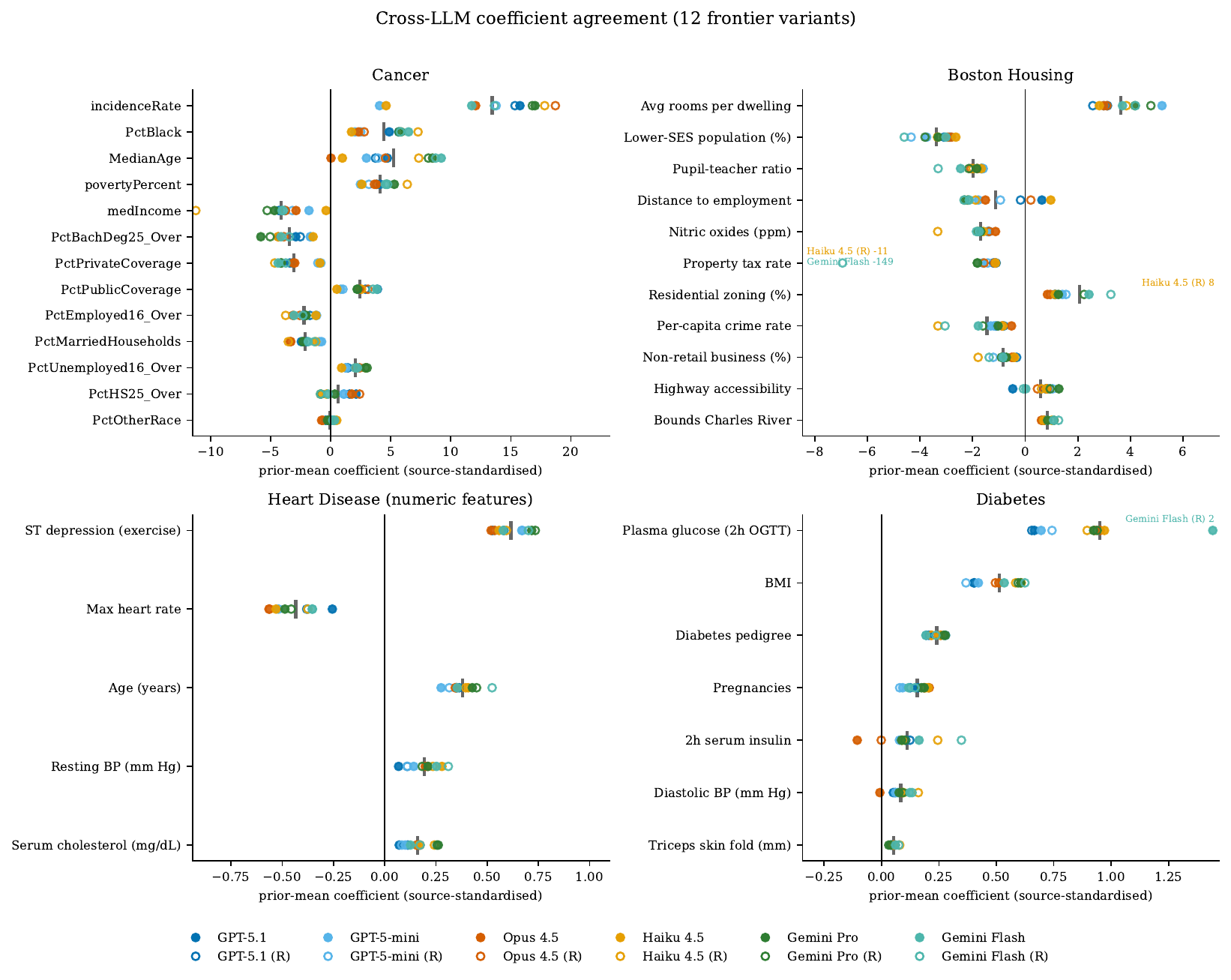}
\caption{Cross-variant coefficient agreement on the four datasets where priors
help. Each panel shows the prior-mean coefficient for all twelve variants per
feature in source-standardised space. Filled circles: non-reasoning. Open
circles: reasoning. Grey tick: across-variant mean. Heart Disease shows numeric
features only. Off-panel points are annotated at the boundary.}
\label{fig:agreement}
\end{figure}

Figure~\ref{fig:agreement} plots the prior-mean coefficient from all twelve variants per feature on the four datasets where priors help. The dominant feature in each panel (cancer incidence rate, property tax rate, exercise-induced ST depression, and plasma glucose) receives the same sign from every variant, with across-variant means clustering tightly. Sign disagreement is concentrated on lower-magnitude, more ambiguous features. Twelve independently prompted variants agreeing on the direction of the dominant predictor of county cancer mortality or of diabetes onset is unlikely to be chance, and is consistent with the priors containing genuine domain structure for the features that matter most.

Agreement on dominant signs does not make provider choice unimportant, as Figure~\ref{fig:selection} shows: smaller magnitude differences, occasional sign disagreements, variance and covariance structure, and individual outlying samples all still matter once a prior enters a fitted model. For Gemini Flash on Boston Housing, $49$ of $50$ samples place the property-tax coefficient near the cluster but a single draw returns roughly $-50$ in raw units (a one-unit tax-rate increase lowering median home value by \$50{,}000) dragging the $M=50$ mean to about $-148$ in standardised space. The CV-tuned $\lambda$ and the selection step contain the damage, and this variant was not deployed on that dataset, but the episode argues for robust prior estimation.

\section{Discussion}

\subsection{When LLM priors help}\label{subsec:whenhelps}

Replacing shrinkage toward zero with shrinkage toward an LLM-elicited target improved out-of-distribution prediction on four of five datasets, and on most of those the improvement is resolved: ten of the twenty-four prior-informed results have a confidence interval for the mean lift that excludes zero in the beneficial direction. Two further results exclude zero on Adult, in the direction of harm and at magnitudes of no practical consequence. The gains are largest in the regression tasks, while in the classification tasks they reside in the probabilistic accuracy rather than ranking. The corresponding prediction intervals in Tables~\ref{tab:regression} and~\ref{tab:classification} are substantially wider and cover zero almost everywhere: the average benefit is measurable, while the benefit of any single deployment is not guaranteed.

The more interesting question is \emph{what} the benefit is, and here the in-domain comparison is decisive. On Boston Housing, Heart Disease and Diabetes the elicited prior improves held-out in-domain accuracy by at least as much as it improves OOD accuracy. On those datasets the prior is simply a better regularisation target than zero: a useful result, but not one about domain shift, and not one that needs an OOD framing to be worth reporting. Cancer is the exception: in-domain the prior is worth $+0.1\%$ and out-of-domain $+3.9\%$. It is the only dataset in the study where the elicited knowledge demonstrably improves transfer rather than fit.

The asymmetry between in-domain improvement for Boston Housing, Heart Disease and Diabetes and OOD improvement for Cancer deserves further attention. The datasets differ in what the source sample can identify. Boston Housing, Heart Disease and Diabetes have $242$ to $316$ training rows against $7$ to $18$ features, such that any sensible shrinkage target helps, both in-domain and out. Cancer has $1{,}106$ rows and $13$ features, thus the source fit is stable in-domain and the prior adds nothing there. Yet it adds several per cent out-of-domain. This is the behaviour of a prior correcting source-specific structure that does not transfer, which is the mechanism the approach aims for. Adult, with $n_{\text{train}} \approx 27{,}000$, shows no effect in either direction: with such a large training sample, the posterior is dominated by the likelihood and the prior is effectively ignored, which from an OOD perspective is not necessarily optimal.

Two consequences follow for how the method should be positioned. It is most clearly useful where source data alone cannot identify stable coefficients, and in that regime its value is mostly better regularisation rather than shift correction. The stronger claim, that elicited knowledge improves transfer beyond what can be achieved by better in-domain regularisation, is supported here by Cancer alone. Testing whether this benefit generalises requires more datasets with enough source data for in-domain gains to be negligible, but with sufficient domain shift for an out-of-domain gain to remain.

\subsection{What selection does, and what it does not}

Depending on the elicitation variant and prior model family, an LLM-elicited prior can either improve or degrade out-of-distribution predictive performance relative to the baseline (Figure~\ref{fig:selection}). The worst candidate performs below baseline on every dataset. An LLM-elicited prior therefore has to be treated as a candidate source of information rather than an authority, and single-provider, single-prompt, single-structure studies are capable of reaching qualitatively different conclusions from the same underlying method. We would encourage reporting the full candidate distribution, not only the deployed configuration.

What source-domain cross-validation does less well is pick between candidates. Measured against the standard error of the paired difference, which is the relevant scale when every candidate is scored on the same folds, the margin between the winner and the best genuinely different runner-up is $0.06$ standard errors on three of the five datasets, $0.25$ on Heart Disease and $1.63$ on Diabetes; on every dataset the selected model is in the middle of the candidate range rather than near the best. The claim the evidence supports is therefore narrower than identification of the best prior: a tuned $\lambda$ together with a selection rule that is at worst uninformative keeps the deployed model away from the worst candidates, and on Adult drives the prior's influence to nothing when it has nothing to offer. The guarantee is safety rather than optimality.

This limitation concerns the source-only selection rule used here, rather than
model selection in general. Cross-validation on source data estimates
source-domain performance, which need not rank candidates by their
target-domain performance. Without information about the target domain, more
extensive tuning on the same source data cannot resolve this mismatch. If
unlabelled target features are available, importance-weighted cross-validation
may provide a more target-relevant criterion under covariate shift, as discussed
in Section~\ref{sec:limitations}. It does not, however, resolve changes in the
relationship between features and outcomes.

\subsection{Evaluating classifiers under label shift}

Our threshold analysis, reported in full in
Appendix~\ref{secA:threshold}, illustrates a risk in comparing
classifiers using $F_1$ at a fixed decision threshold. In our
Heart Disease and Diabetes splits, the positive-class prevalence
is higher in the OOD population than in the source population.
At a threshold of $0.5$, differences in how many observations
each model classifies as positive can therefore create an
apparent $F_1$ advantage that disappears when the baseline is
also allowed to select its threshold on source data. In addition to the threshold-free metrics of Table~\ref{tab:classification}, we evaluated every classifier with $F_{1}$, and for threshold-dependent metrics of that kind the effect is not subtle: on Diabetes it accounts for an apparent $F_1$ improvement of $+0.089$ with a confidence interval comfortably excluding zero, which disappears entirely when the baseline is allowed the same operating-point freedom. AUC is invariant to label shift and was correctly reported as unchanged; the threshold-dependent metrics were not.

Three recommendations follow: report proper scoring rules, which are threshold-free and reward exactly the probability quality that prior-informed shrinkage delivers; if threshold-dependent metrics are reported, tune the threshold for every model including baselines, on source data only; and document prevalence shift as a property of the design rather than an incidental detail.

\subsection{Limitations and future work}\label{sec:limitations}

A major threat to validity is possible overlap between the LLMs' training
corpora and the public datasets used here. Boston Housing, the UCI Heart
Disease cohorts, Pima Diabetes and Adult are long-standing benchmarks, but
we cannot determine whether, or in what form, any queried model encountered
them during training. Prior work documents memorisation and contamination
risks in tabular benchmarks
\cite{bordt2024elephants,silvestri2025contamination,gorla2026illusion}.
This matters for interpreting the elicited coefficients: when an LLM gives
nearly identical answers across samples, we cannot tell whether it is
expressing stable domain knowledge or recalling a familiar dataset.
Evaluation on datasets unlikely to have appeared in model training would
provide a stronger test.

A separate concern applies to Boston Housing. That dataset has been withdrawn
from major software distributions because one of its features encodes a
racial-segregation hypothesis in its functional form. We remove the feature in
question and fit the remaining eleven, as described in the Methods. We keep the
dataset itself because it is a long-standing benchmark in this literature and
including it makes our results comparable with earlier work, but we give its
coefficients no causal interpretation. The conclusions do not rest on it. The
claim that elicited knowledge supplies information the source domain cannot is
supported by Cancer, and it would stand unchanged if Boston Housing were
excluded.

Our comparisons also have a deliberately narrow scope. The matched
baselines shrink coefficients toward zero, using ridge regression for the
regression tasks and $L_2$-regularised logistic regression for
classification. They isolate the question of whether the elicited prior
improves on zero-centred shrinkage within the same modelling framework;
they do not establish that this approach outperforms other ways of using
LLM knowledge. A direct comparison with AutoElicit or LoID would require
new elicitation queries, because their prompting and extraction procedures
cannot be reconstructed from our stored coefficient draws. We did not
undertake that comparison here. Our stored draws allow the present analysis
to be reproduced offline, whereas new queries to commercial models could
introduce differences due to model updates unless their versions were
recorded.

Each dataset contributes only one source--target split. Consequently, a
gain on its OOD partition need not generalise to other plausible shifts
within the same domain. The uncertainty intervals account for repeated
source-data splits, re-tuning of $\lambda$, and resampling of the observed
target partition, but hold the selected LLM variant and its elicited draws
fixed. Repeating variant selection within each source split would assess
the stability of that selection step; independently repeating elicitation
would be needed to assess variability in the priors themselves. The small
OOD partitions on several datasets also leave considerable uncertainty
about the benefit of a single deployment. Larger target samples and
multiple distinct domain splits would strengthen these conclusions.

A related interpretive limitation is that Boston Housing, Heart Disease
and Diabetes show in-domain gains at least as large as their OOD gains.
For those datasets, the evidence points primarily to better regularisation,
not to an effect specific to domain shift. Cancer is the only dataset here
with a substantial OOD gain and negligible in-domain gain. More datasets
with sufficient source data to make ordinary in-domain regularisation gains
small, but with meaningful domain shift, would help test whether this
pattern generalises.

The elicitation prompts also include author-written hints about the
expected signs of dominant features (Appendix~\ref{secA:prompt}). The
placebo tests show that feature assignments and signs matter for
predictive performance, but they cannot establish that the useful
directions originated with the LLM rather than with these hints. The
elicited magnitudes and uncertainty structure remain relevant, particularly
on Boston Housing, where the prior families that use elicited spread
outperform Prior-Mean. An otherwise identical elicitation without the
hint block would clarify what the LLM contributes independently.

A further limitation concerns model selection under domain shift. In
Eq.~\ref{eq:map}, the model is fitted to source-domain observations, and
$\lambda$ is selected by source-domain cross-validation. That procedure
estimates source performance, which need not rank candidates by their
target performance. Adult illustrates the ambiguity: cross-validation
assigns the prior very little weight, but this alone cannot tell us
whether a stronger prior would have transferred better. With no
information about the target domain, more extensive tuning on the same
source data cannot resolve that uncertainty.

If unlabelled target features are available and the conditional
relationship between features and outcomes is sufficiently stable,
importance-weighted validation could offer a more target-relevant
selection criterion \cite{shimodaira2000covariate,sugiyama2007iwcv}.
This is a different setting from the source-only selection studied here:
it requires target covariates, adequate overlap between domains, and
assumptions that may fail under concept shift. Likelihood tempering
\cite{grunwald2017inconsistency,deheide2020safe} and alternative scaling
of prior influence \cite{ibrahim2000power} address a related but distinct
question: how strongly should source observations constrain a model
relative to external information? Neither, on its own, identifies the
weight that minimises target risk without target information or additional
assumptions. These approaches are therefore directions for future study,
rather than remedies established by the present results.

Finally, several extensions concern the priors themselves. Combining
repeated coefficient sampling with AutoElicit-style prompt rephrasing
could capture both within-prompt and cross-prompt variability. Robust
estimation of $\muhat$, or pooling samples across providers, could reduce
sensitivity to outlying coefficient draws. Covariance shrinkage could
also stabilise Correlated-Prior, whose covariance matrix is estimated
from $50$ samples. A comparison of these approaches and other
elicitation methods on common datasets and domain splits would help
separate the value of the prior information from the way it is extracted
and encoded.
\subsection{Conclusion}

LLM-elicited coefficient priors used as structured shrinkage in regularised regression improve OOD prediction on four of five real domain splits. For ten of the twenty-four prior-informed evaluations, the mean improvement is statistically significant at the standard 95\% level. A placebo control attributes the performance lift over baseline to the priors' feature-specific content rather than their scale. The effect is conditional, and largest when training data are limited relative to the number of features and absent where the training data are already sufficient. However, prediction intervals for a single deployment remain wide and frequently cross zero due to variability in training splits and test-sample evaluation. Consequently, while the average benefit across repeated trials is statistically established, an improvement for any single run cannot be guaranteed.

In addition to predictive performance, two broader methodological findings emerge. First, the choice of language-model variant and prior structure matters: some combinations improve out-of-distribution predictive performance, while others degrade it relative to the baseline. Because source-domain cross-validation cannot reliably rank these candidates, the practical guarantee is protection against the worst-performing configurations rather than the identification of the optimal one; transparently reporting the full candidate distribution should therefore become standard practice. Second, evaluating classifiers at a fixed decision threshold—such as the common default of $F_1$ at a 0.5 cut-off—substantially inflates apparent performance gains whenever the domain split alters the class distribution. In such settings, threshold-free proper scoring rules provide a much more reliable basis for comparison.

Finally, the central open question concerns whether language model priors act as true domain shift-adaptation mechanisms or merely as variance regularisers. Across most evaluated datasets, the elicited prior improves in-domain accuracy just as much as out-of-domain accuracy, indicating that it functions primarily as an effective regularisation target to counter small-sample noise rather than as a targeted remedy for domain shift. Only on Cancer, where the training sample is large enough to render in-domain regularisation gains negligible, does the prior provide an improvement specific to out-of-distribution prediction.

\subsection*{Data availability}

All five datasets, in the exact form used here, are in the repository below. Four are available from their original distributors: Boston Housing \citep{harrison1978hedonic}, the UCI Heart Disease database \citep{detrano1989heart}, the Pima Indians Diabetes Database \citep{smith1988pima} and the UCI Adult/Census Income dataset \citep{kohavi1996adult}, the latter two under CC BY 4.0. The county-level cancer mortality compilation \citep{rippner2017cancer} is a special case: it was published openly on data.world, but that platform has since retired its open data community and the original page no longer resolves, so our copy is the only one we can point to. Its underlying sources: the US Census American Community Survey, Cancer.gov and ClinicalTrials.gov, are US federal government outputs and are not subject to copyright; we redistribute the compilation to keep the analysis reproducible, with attribution to its compiler, and will remove it on request.

\subsection*{Code availability}

The elicitation and modelling pipeline, the $50$ validated coefficient samples for every (dataset, provider, mode) combination, the scripts for all follow-up analyses, and the result files and run logs behind every table and figure are available at \url{https://github.com/hugohammer/llm-priors-ood}. Because the elicited samples are stored, every result in this paper can be reproduced without any further LLM API calls.

\subsection*{Author contributions}

H.L.H.\ conceived the project. S.L.\ and H.L.H.\ implemented the study and drafted the manuscript. K.H.H.\ and H.L.H.\ supervised S.L.'s work on the project. K.H.H.\ reviewed and critically revised the manuscript at several stages during its preparation. All authors read and approved the final manuscript.

\subsection*{Competing interests}

The authors declare no competing interests.

\bibliography{references}

\begin{thebibliography}{10}
\expandafter\ifx\csname url\endcsname\relax
  \def\url#1{\burl{#1}}\fi
\expandafter\ifx\csname urlprefix\endcsname\relax\def\urlprefix{URL }\fi
\providecommand{\bibinfo}[2]{#2}
\providecommand{\eprint}[2][]{\url{#2}}
\providecommand{\doi}[1]{\url{https://doi.org/#1}}
\bibcommenthead

\bibitem{liu2021oodsurvey}
\bibinfo{author}{Liu, J.} \emph{et~al.}
\newblock \bibinfo{title}{Towards out-of-distribution generalization: A survey}
  (\bibinfo{year}{2021}).
\newblock
  \bibinfo{eprint}{{\href{https://arxiv.org/abs/2108.13624}{{arXiv:2108.13624}}}}.

\bibitem{quinonerocandela2009shift}
\bibinfo{editor}{Qui{\~n}onero-Candela, J.}, \bibinfo{editor}{Sugiyama, M.},
  \bibinfo{editor}{Schwaighofer, A.} \& \bibinfo{editor}{Lawrence, N.~D.} (eds)
  \emph{\bibinfo{title}{Dataset Shift in Machine Learning}}
  (\bibinfo{publisher}{MIT Press}, \bibinfo{address}{Cambridge, MA},
  \bibinfo{year}{2009}).

\bibitem{morenotorres2012unifying}
\bibinfo{author}{Moreno-Torres, J.~G.}, \bibinfo{author}{Raeder, T.},
  \bibinfo{author}{Alaiz-Rodr{\'i}guez, R.}, \bibinfo{author}{Chawla, N.~V.} \&
  \bibinfo{author}{Herrera, F.}
\newblock \bibinfo{title}{A unifying view on dataset shift in classification}.
\newblock \emph{\bibinfo{journal}{Pattern Recognition}}
  \textbf{\bibinfo{volume}{45}}, \bibinfo{pages}{521--530}
  (\bibinfo{year}{2012}).

\bibitem{lakshminarayanan2017deepensembles}
\bibinfo{author}{Lakshminarayanan, B.}, \bibinfo{author}{Pritzel, A.} \&
  \bibinfo{author}{Blundell, C.}
\newblock \bibinfo{title}{Simple and scalable predictive uncertainty estimation
  using deep ensembles}.
\newblock \emph{\bibinfo{journal}{Advances in Neural Information Processing
  Systems 30 (NIPS 2017)}}  (\bibinfo{year}{2017}).

\bibitem{purucker2026beyondiid}
\bibinfo{author}{Purucker, L.} \emph{et~al.}
\newblock \bibinfo{title}{Beyond iid: How general are tabular foundation
  models, really?} (\bibinfo{year}{2026}).
\newblock \bibinfo{note}{ArXiv:2606.30410}.

\bibitem{sokolov2016pathway}
\bibinfo{author}{Sokolov, A.}, \bibinfo{author}{Carlin, D.~E.},
  \bibinfo{author}{Paull, E.~O.}, \bibinfo{author}{Baertsch, R.} \&
  \bibinfo{author}{Stuart, J.~M.}
\newblock \bibinfo{title}{Pathway-based genomics prediction using generalized
  elastic net}.
\newblock \emph{\bibinfo{journal}{PLOS Computational Biology}}
  \textbf{\bibinfo{volume}{12}}, \bibinfo{pages}{1--23} (\bibinfo{year}{2016}).
\newblock \urlprefix\url{https://doi.org/10.1371/journal.pcbi.1004790}.

\bibitem{ohagan2019elicitation}
\bibinfo{author}{O'Hagan, A.}
\newblock \bibinfo{title}{Expert knowledge elicitation: Subjective but
  scientific}.
\newblock \emph{\bibinfo{journal}{The American Statistician}}
  \textbf{\bibinfo{volume}{73}}, \bibinfo{pages}{69--81}
  (\bibinfo{year}{2019}).

\bibitem{vandeschoot2021bayesian}
\bibinfo{author}{van~de Schoot, R.} \emph{et~al.}
\newblock \bibinfo{title}{Bayesian statistics and modelling}.
\newblock \emph{\bibinfo{journal}{Nature Reviews Methods Primers}}
  \textbf{\bibinfo{volume}{1}} (\bibinfo{year}{2021}).

\bibitem{bishop2006prml}
\bibinfo{author}{Bishop, C.~M.}
\newblock \emph{\bibinfo{title}{Pattern Recognition and Machine Learning}}
  (\bibinfo{publisher}{Springer}, \bibinfo{year}{2006}).

\bibitem{hastie2009elements}
\bibinfo{author}{Hastie, T.}, \bibinfo{author}{Tibshirani, R.} \&
  \bibinfo{author}{Friedman, J.}
\newblock \emph{\bibinfo{title}{The Elements of Statistical Learning: Data
  Mining, Inference, and Prediction}} \bibinfo{edition}{2nd} edn
  (\bibinfo{publisher}{Springer}, \bibinfo{year}{2009}).

\bibitem{hoerl1970ridge}
\bibinfo{author}{Hoerl, A.~E.} \& \bibinfo{author}{Kennard, R.~W.}
\newblock \bibinfo{title}{Ridge regression: Biased estimation for nonorthogonal
  problems}.
\newblock \emph{\bibinfo{journal}{Technometrics}}
  \textbf{\bibinfo{volume}{12}}, \bibinfo{pages}{55--67}
  (\bibinfo{year}{1970}).

\bibitem{brown2020gpt3}
\bibinfo{author}{Brown, T.~B.} \emph{et~al.}
\newblock \bibinfo{title}{Language models are few-shot learners}.
\newblock \emph{\bibinfo{journal}{Advances in Neural Information Processing
  Systems}} \textbf{\bibinfo{volume}{33}} (\bibinfo{year}{2020}).

\bibitem{bommasani2021foundation}
\bibinfo{author}{Bommasani, R.} \emph{et~al.}
\newblock \bibinfo{title}{On the opportunities and risks of foundation models}
  (\bibinfo{year}{2021}).
\newblock
  \bibinfo{eprint}{{\href{https://arxiv.org/abs/2108.07258}{{arXiv:2108.07258}}}}.

\bibitem{minaee2024llmsurvey}
\bibinfo{author}{Minaee, S.} \emph{et~al.}
\newblock \bibinfo{title}{Large language models: A survey}
  (\bibinfo{year}{2024}).
\newblock
  \bibinfo{eprint}{{\href{https://arxiv.org/abs/2402.06196}{{arXiv:2402.06196}}}}.

\bibitem{goh2024llmdiagnostic}
\bibinfo{author}{Goh, E.}, \bibinfo{author}{Gallo, R.}, \bibinfo{author}{Hom,
  J.} \emph{et~al.}
\newblock \bibinfo{title}{Large language model influence on diagnostic
  reasoning: A randomized clinical trial}.
\newblock \emph{\bibinfo{journal}{JAMA Network Open}}
  \textbf{\bibinfo{volume}{7}}, \bibinfo{pages}{e2440969}
  (\bibinfo{year}{2024}).

\bibitem{evans2025survival}
\bibinfo{author}{Evans, R.}, \bibinfo{author}{Felland, M.},
  \bibinfo{author}{Evans, S.} \& \bibinfo{author}{Sloan, L.}
\newblock \bibinfo{title}{Large language model-derived priors can improve
  bayesian survival analyses: A glioblastoma application}
  (\bibinfo{year}{2025}).
\newblock \bibinfo{note}{ArXiv:2511.01778}.

\bibitem{bender2021parrots}
\bibinfo{author}{Bender, E.~M.}, \bibinfo{author}{Gebru, T.},
  \bibinfo{author}{McMillan-Major, A.} \& \bibinfo{author}{Shmitchell, S.}
\newblock \bibinfo{title}{On the dangers of stochastic parrots: Can language
  models be too big?}
\newblock \emph{\bibinfo{journal}{Proceedings of the 2021 ACM Conference on
  Fairness, Accountability, and Transparency}} \bibinfo{pages}{610--623}
  (\bibinfo{year}{2021}).

\bibitem{liu2023prompt}
\bibinfo{author}{Liu, P.} \emph{et~al.}
\newblock \bibinfo{title}{Pre-train, prompt, and predict: A systematic survey
  of prompting methods in natural language processing}.
\newblock \emph{\bibinfo{journal}{ACM Computing Surveys}}
  \textbf{\bibinfo{volume}{55}}, \bibinfo{pages}{Article 195}
  (\bibinfo{year}{2023}).

\bibitem{gallegos2024bias}
\bibinfo{author}{Gallegos, I.~O.} \emph{et~al.}
\newblock \bibinfo{title}{Bias and fairness in large language models: A
  survey}.
\newblock \emph{\bibinfo{journal}{Computational Linguistics}}
  \textbf{\bibinfo{volume}{50}}, \bibinfo{pages}{1097--1179}
  (\bibinfo{year}{2024}).

\bibitem{huang2025hallucination}
\bibinfo{author}{Huang, L.} \emph{et~al.}
\newblock \bibinfo{title}{A survey on hallucination in large language models:
  Principles, taxonomy, challenges, and open questions}.
\newblock \emph{\bibinfo{journal}{ACM Transactions on Information Systems}}
  \textbf{\bibinfo{volume}{43}}, \bibinfo{pages}{Article 73}
  (\bibinfo{year}{2025}).

\bibitem{gneiting2007scoring}
\bibinfo{author}{Gneiting, T.} \& \bibinfo{author}{Raftery, A.~E.}
\newblock \bibinfo{title}{Strictly proper scoring rules, prediction, and
  estimation}.
\newblock \emph{\bibinfo{journal}{Journal of the American Statistical
  Association}} \textbf{\bibinfo{volume}{102}}, \bibinfo{pages}{359--378}
  (\bibinfo{year}{2007}).

\bibitem{riegler2025priors}
\bibinfo{author}{Riegler, M.~A.}, \bibinfo{author}{Hellton, K.~H.},
  \bibinfo{author}{Thambawita, V.} \& \bibinfo{author}{Hammer, H.~L.}
\newblock \bibinfo{title}{Using large language models to suggest informative
  prior distributions in {Bayesian} regression analysis}.
\newblock \emph{\bibinfo{journal}{Scientific Reports}}
  \textbf{\bibinfo{volume}{15}}, \bibinfo{pages}{33386} (\bibinfo{year}{2025}).

\bibitem{capstick2025autoelicit}
\bibinfo{author}{Capstick, A.}, \bibinfo{author}{Krishnan, R.~G.} \&
  \bibinfo{author}{Barnaghi, P.}
\newblock \bibinfo{title}{{AutoElicit}: Using large language models for expert
  prior elicitation in predictive modelling}.
\newblock \emph{\bibinfo{journal}{Proceedings of the 42nd International
  Conference on Machine Learning}} \textbf{\bibinfo{volume}{267}},
  \bibinfo{pages}{6746--6777} (\bibinfo{year}{2025}).

\bibitem{huang2025llmprior}
\bibinfo{author}{Huang, Y.}
\newblock \bibinfo{title}{{LLM-Prior}: A framework for knowledge-driven prior
  elicitation and aggregation} (\bibinfo{year}{2025}).
\newblock \bibinfo{note}{ArXiv:2508.03766}.

\bibitem{zhang2025llmlasso}
\bibinfo{author}{Zhang, E.} \emph{et~al.}
\newblock \bibinfo{title}{{LLM-Lasso}: A robust framework for domain-informed
  feature selection and regularization} (\bibinfo{year}{2025}).
\newblock
  \bibinfo{eprint}{{\href{https://arxiv.org/abs/2502.10648}{{arXiv:2502.10648}}}}.

\bibitem{requeima2024llmprocesses}
\bibinfo{author}{Requeima, J.}, \bibinfo{author}{Bronskill, J.},
  \bibinfo{author}{Choi, D.}, \bibinfo{author}{Turner, R.~E.} \&
  \bibinfo{author}{Duvenaud, D.}
\newblock \bibinfo{title}{Llm processes: Numerical predictive distributions
  conditioned on natural language} (\bibinfo{year}{2024}).
\newblock \urlprefix\url{https://arxiv.org/abs/2405.12856}.
\newblock
  \bibinfo{eprint}{{\href{https://arxiv.org/abs/2405.12856}{{arXiv:2405.12856}}}}.

\bibitem{zeng2024llmembeddingsimprovetesttime}
\bibinfo{author}{Zeng, Y.}, \bibinfo{author}{Liu, J.}, \bibinfo{author}{Lam,
  H.} \& \bibinfo{author}{Namkoong, H.}
\newblock \bibinfo{title}{Llm embeddings improve test-time adaptation to
  tabular $y|x$-shifts} (\bibinfo{year}{2024}).
\newblock \urlprefix\url{https://arxiv.org/abs/2410.07395}.
\newblock
  \bibinfo{eprint}{{\href{https://arxiv.org/abs/2410.07395}{{arXiv:2410.07395}}}}.

\bibitem{amin2026integratingmachinelearningensembles}
\bibinfo{author}{Amin, M.~T.} \emph{et~al.}
\newblock \bibinfo{title}{Integrating machine learning ensembles and large
  language models for heart disease prediction using voting fusion}
  (\bibinfo{year}{2026}).
\newblock \urlprefix\url{https://arxiv.org/abs/2602.22280}.
\newblock
  \bibinfo{eprint}{{\href{https://arxiv.org/abs/2602.22280}{{arXiv:2602.22280}}}}.

\bibitem{rezaeimanesh2026languagemodelsknowdont}
\bibinfo{author}{Rezaeimanesh, S.}, \bibinfo{author}{Thind, K.},
  \bibinfo{author}{Siddiqui, F.} \& \bibinfo{author}{Ghassemi, M.~M.}
\newblock \bibinfo{title}{What language models know but don't say:
  Non-generative prior extraction for generalization} (\bibinfo{year}{2026}).
\newblock \urlprefix\url{https://arxiv.org/abs/2601.17609}.
\newblock \bibinfo{note}{ArXiv:2601.17609v3},
  \bibinfo{eprint}{{\href{https://arxiv.org/abs/2601.17609}{{arXiv:2601.17609}}}}.

\bibitem{harrison1978hedonic}
\bibinfo{author}{Harrison, D.} \& \bibinfo{author}{Rubinfeld, D.~L.}
\newblock \bibinfo{title}{Hedonic housing prices and the demand for clean air}.
\newblock \emph{\bibinfo{journal}{Journal of Environmental Economics and
  Management}} \textbf{\bibinfo{volume}{5}}, \bibinfo{pages}{81--102}
  (\bibinfo{year}{1978}).

\bibitem{rippner2017cancer}
\bibinfo{author}{Rippner, N.}
\newblock \bibinfo{title}{{OLS} regression challenge: County-level cancer
  mortality}.
\newblock \bibinfo{howpublished}{data.world (platform retired); archived copy
  at \url{https://github.com/hugohammer/llm-priors-ood}}
  (\bibinfo{year}{2017}).
\newblock \bibinfo{note}{County-level compilation (\texttt{cancer\_reg.csv},
  3{,}047 counties) aggregated from the US Census American Community Survey,
  ClinicalTrials.gov and Cancer.gov. The original page,
  data.world/nrippner/ols-regression-challenge, no longer resolves following
  the retirement of the data.world open data community}.

\bibitem{detrano1989heart}
\bibinfo{author}{Detrano, R.} \emph{et~al.}
\newblock \bibinfo{title}{International application of a new probability
  algorithm for the diagnosis of coronary artery disease}.
\newblock \emph{\bibinfo{journal}{The American Journal of Cardiology}}
  \textbf{\bibinfo{volume}{64}}, \bibinfo{pages}{304--310}
  (\bibinfo{year}{1989}).

\bibitem{smith1988pima}
\bibinfo{author}{Smith, J.~W.}, \bibinfo{author}{Everhart, J.~E.},
  \bibinfo{author}{Dickson, W.~C.}, \bibinfo{author}{Knowler, W.~C.} \&
  \bibinfo{author}{Johannes, R.~S.}
\newblock \bibinfo{title}{Using the {ADAP} learning algorithm to forecast the
  onset of diabetes mellitus}.
\newblock \emph{\bibinfo{journal}{Proceedings of the Annual Symposium on
  Computer Application in Medical Care}} \bibinfo{pages}{261--265}
  (\bibinfo{year}{1988}).

\bibitem{kohavi1996adult}
\bibinfo{author}{Kohavi, R.}
\newblock \bibinfo{title}{Scaling up the accuracy of naive-{Bayes} classifiers:
  A decision-tree hybrid}.
\newblock \emph{\bibinfo{journal}{Proceedings of the Second International
  Conference on Knowledge Discovery and Data Mining (KDD)}}
  \bibinfo{pages}{202--207} (\bibinfo{year}{1996}).

\bibitem{pedregosa2011sklearn}
\bibinfo{author}{Pedregosa, F.} \emph{et~al.}
\newblock \bibinfo{title}{Scikit-learn: Machine learning in {Python}}.
\newblock \emph{\bibinfo{journal}{Journal of Machine Learning Research}}
  \textbf{\bibinfo{volume}{12}}, \bibinfo{pages}{2825--2830}
  (\bibinfo{year}{2011}).
\newblock \bibinfo{note}{The \texttt{load\_boston} loader was deprecated in
  version 1.0 and removed in 1.2 over the ethical problem with the B variable.}

\bibitem{wang2017oilseeds}
\bibinfo{author}{Wang, S.}, \bibinfo{author}{Kreuzer, M.},
  \bibinfo{author}{Braun, U.} \& \bibinfo{author}{Schwarm, A.}
\newblock \bibinfo{title}{Effect of unconventional oilseeds (safflower, poppy,
  hemp, camelina) on in vitro ruminal methane production and fermentation}.
\newblock \emph{\bibinfo{journal}{Journal of the Science of Food and
  Agriculture}} \textbf{\bibinfo{volume}{97}}, \bibinfo{pages}{3864--3870}
  (\bibinfo{year}{2017}).

\bibitem{bordt2024elephants}
\bibinfo{author}{Bordt, S.}, \bibinfo{author}{Nori, H.},
  \bibinfo{author}{Rodrigues, V.}, \bibinfo{author}{Nushi, B.} \&
  \bibinfo{author}{Caruana, R.}
\newblock \bibinfo{title}{Elephants never forget: Memorization and learning of
  tabular data in large language models}.
\newblock \emph{\bibinfo{journal}{Conference on Language Modeling (COLM)}}
  (\bibinfo{year}{2024}).
\newblock \bibinfo{note}{ArXiv:2404.06209}.

\bibitem{silvestri2025contamination}
\bibinfo{author}{Silvestri, M.}, \bibinfo{author}{Veglianti, F.},
  \bibinfo{author}{Giorgi, F.}, \bibinfo{author}{Silvestri, F.} \&
  \bibinfo{author}{Tolomei, G.}
\newblock \bibinfo{title}{When large language models know the table: A
  framework for assessing data contamination in tabular datasets}
  (\bibinfo{year}{2025}).
\newblock \bibinfo{note}{ArXiv:2510.20351}.

\bibitem{gorla2026illusion}
\bibinfo{author}{Gorla, A.} \& \bibinfo{author}{Puduppully, R.}
\newblock \bibinfo{title}{The illusion of generalization in tabular language
  models} (\bibinfo{year}{2026}).
\newblock \bibinfo{note}{ArXiv:2602.04031}.

\bibitem{shimodaira2000covariate}
\bibinfo{author}{Shimodaira, H.}
\newblock \bibinfo{title}{Improving predictive inference under covariate shift
  by weighting the log-likelihood function}.
\newblock \emph{\bibinfo{journal}{Journal of Statistical Planning and
  Inference}} \textbf{\bibinfo{volume}{90}}, \bibinfo{pages}{227--244}
  (\bibinfo{year}{2000}).

\bibitem{sugiyama2007iwcv}
\bibinfo{author}{Sugiyama, M.}, \bibinfo{author}{Krauledat, M.} \&
  \bibinfo{author}{M{\"u}ller, K.-R.}
\newblock \bibinfo{title}{Covariate shift adaptation by importance weighted
  cross validation}.
\newblock \emph{\bibinfo{journal}{Journal of Machine Learning Research}}
  \textbf{\bibinfo{volume}{8}}, \bibinfo{pages}{985--1005}
  (\bibinfo{year}{2007}).

\bibitem{grunwald2017inconsistency}
\bibinfo{author}{Gr{\"u}nwald, P.} \& \bibinfo{author}{van Ommen, T.}
\newblock \bibinfo{title}{Inconsistency of {Bayesian} inference for
  misspecified linear models, and a proposal for repairing it}.
\newblock \emph{\bibinfo{journal}{Bayesian Analysis}}
  \textbf{\bibinfo{volume}{12}}, \bibinfo{pages}{1069--1103}
  (\bibinfo{year}{2017}).

\bibitem{deheide2020safe}
\bibinfo{author}{de~Heide, R.}, \bibinfo{author}{Kirichenko, A.},
  \bibinfo{author}{Gr{\"u}nwald, P.} \& \bibinfo{author}{Mehta, N.~A.}
\newblock \bibinfo{title}{Safe-{Bayesian} generalized linear regression}.
\newblock \emph{\bibinfo{journal}{Proceedings of the 23rd International
  Conference on Artificial Intelligence and Statistics (AISTATS), PMLR}}
  \textbf{\bibinfo{volume}{108}}, \bibinfo{pages}{2623--2633}
  (\bibinfo{year}{2020}).

\bibitem{ibrahim2000power}
\bibinfo{author}{Ibrahim, J.~G.} \& \bibinfo{author}{Chen, M.-H.}
\newblock \bibinfo{title}{Power prior distributions for regression models}.
\newblock \emph{\bibinfo{journal}{Statistical Science}}
  \textbf{\bibinfo{volume}{15}}, \bibinfo{pages}{46--60}
  (\bibinfo{year}{2000}).

\bibitem{saerens2002adjusting}
\bibinfo{author}{Saerens, M.}, \bibinfo{author}{Latinne, P.} \&
  \bibinfo{author}{Decaestecker, C.}
\newblock \bibinfo{title}{Adjusting the outputs of a classifier to new a priori
  probabilities: A simple procedure}.
\newblock \emph{\bibinfo{journal}{Neural Computation}}
  \textbf{\bibinfo{volume}{14}}, \bibinfo{pages}{21--41}
  (\bibinfo{year}{2002}).

\end{thebibliography}

\clearpage

\begin{appendices}

\section{Supplementary figures}\label{secA1}

\renewcommand{\thefigure}{S\arabic{figure}}
\setcounter{figure}{0}
\renewcommand{\thetable}{S\arabic{table}}
\setcounter{table}{0}

\begin{figure}[htbp]
\centering
\includegraphics[width=\linewidth]{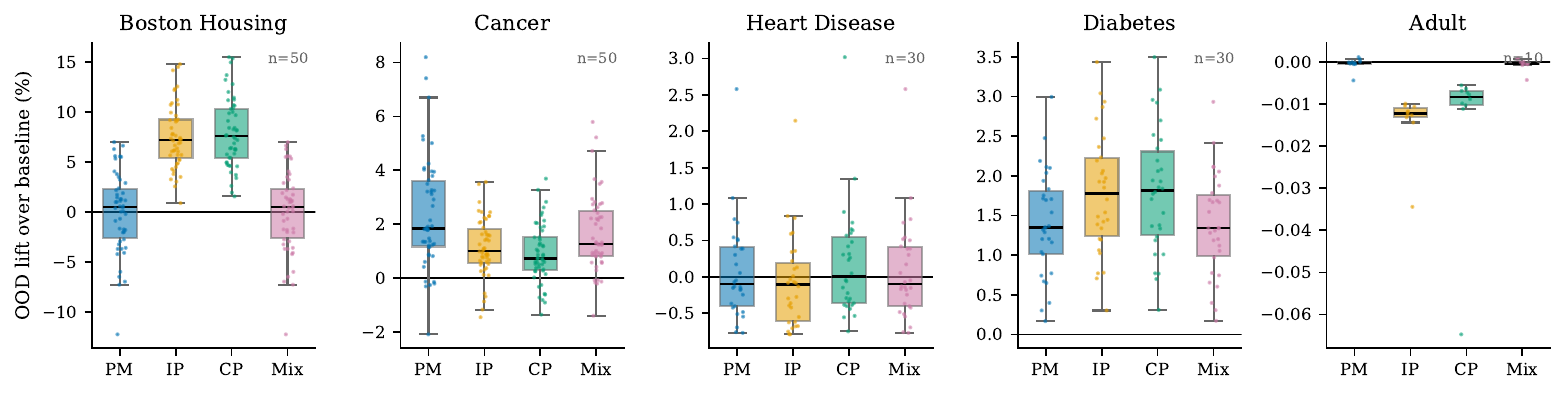}
\caption{OOD lift over baseline across ID resplits, with $\lambda$ re-tuned on
every resplit. Boxes show the interquartile range and median; points are
individual resplits. Lift is MSE reduction for Boston Housing and Cancer and
AUC gain for the classification datasets. The single-split values reported in
Table~\ref{tab:regression} are in the upper part of these distributions.}
\label{fig:resplits}
\end{figure}

\begin{figure}[htbp]
\centering
\includegraphics[width=0.45\linewidth]{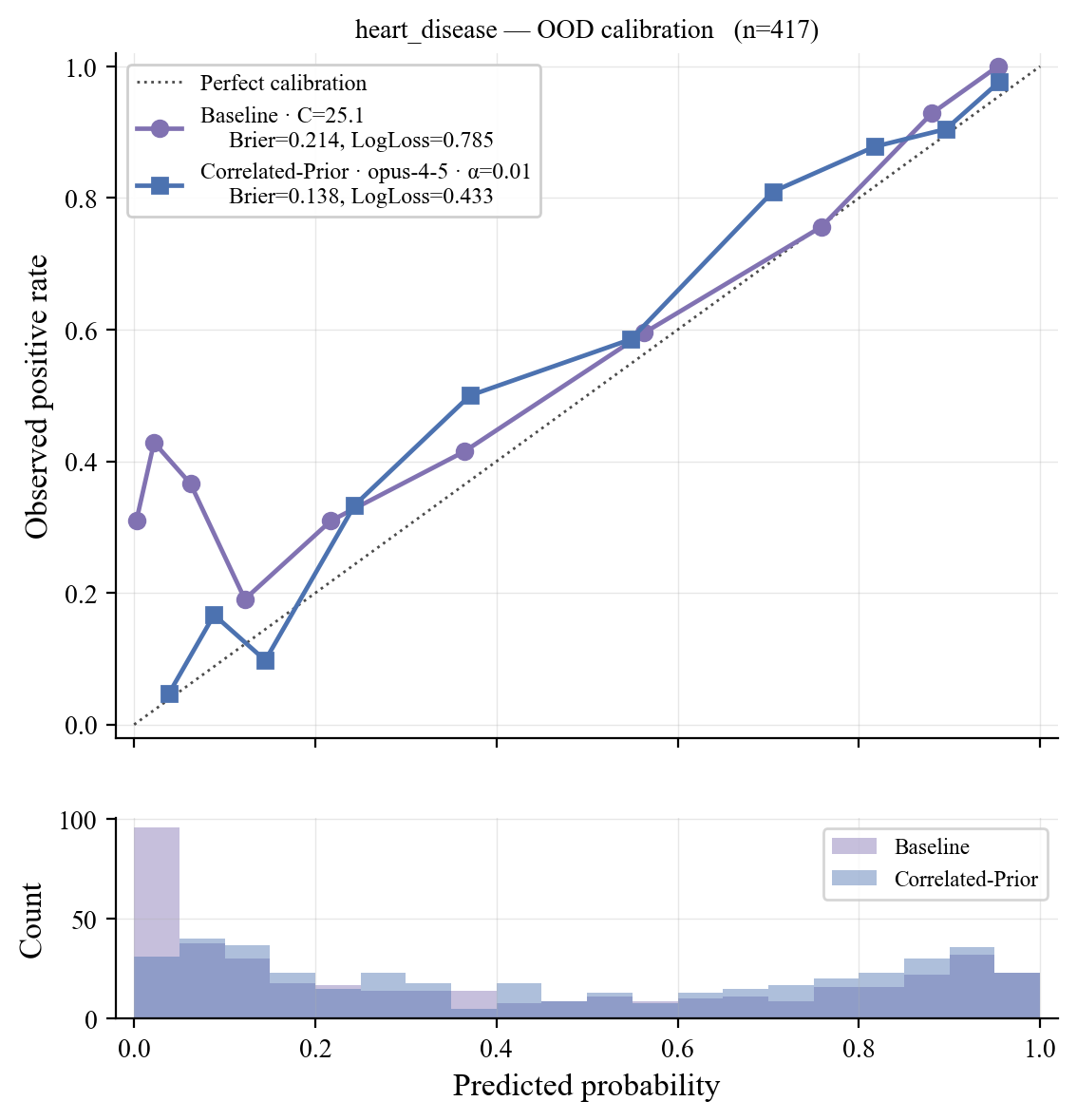}\hfill
\includegraphics[width=0.45\linewidth]{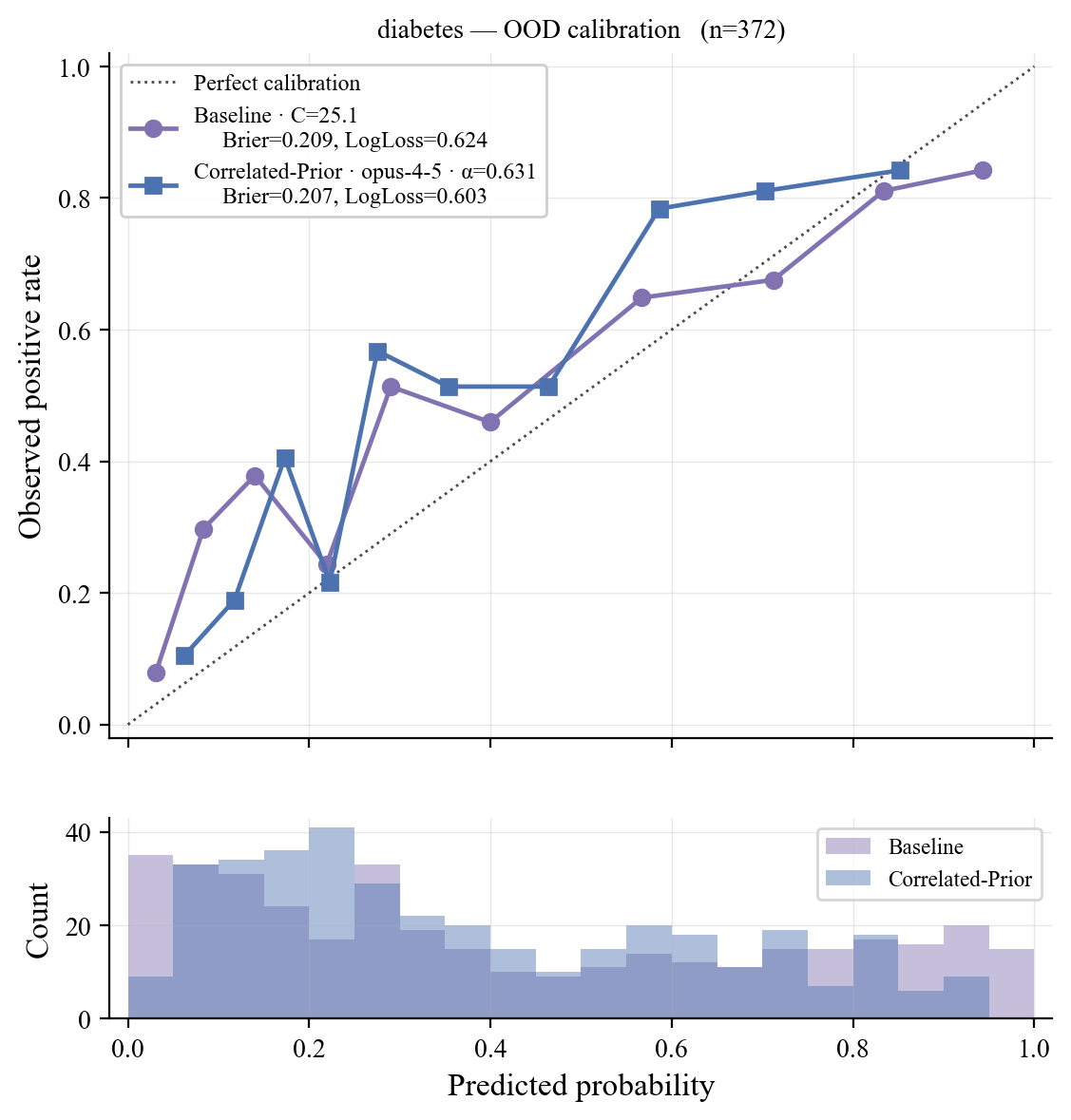}
\caption{OOD calibration on Heart Disease (left) and Diabetes (right).
Top panels: 10-quantile reliability curves for the deployed baseline and the
deployed best-prior family, with the dotted diagonal marking perfect
calibration. Bottom panels: histograms of predicted probabilities. AUC is
essentially unchanged in both cases while Brier and log loss improve.}
\label{fig:calibration}
\end{figure}

\begin{table}[htbp]
\centering
\footnotesize
\caption{\emph{Supplementary.} OOD lift over baseline across ID resplits ($n=50$ for Boston Housing
and Cancer, $30$ for Heart Disease and Diabetes, $10$ for Adult).
``Single split'' is the value from Tables~\ref{tab:regression} and~\ref{tab:classification}. ``Below'' is the fraction of resplits in which the
prior-informed model is worse than baseline.}
\label{tab:resplits}
\begin{tabular}{@{}llrrrrr@{}}
\toprule
Dataset & Family & Single split & Mean & SD & [2.5, 97.5] & Below \\
\midrule
Boston Housing & Prior-Mean        & $+2.7\%$  & $-0.02\%$ & 4.02 & [$-7.22$, $+6.55$]  & 42\% \\
Boston Housing & Independent-Prior & $+10.3\%$ & $+7.52\%$ & 3.20 & [$+2.66$, $+14.43$] & 0\% \\
Boston Housing & Correlated-Prior  & $+11.2\%$ & $+7.94\%$ & 3.48 & [$+2.09$, $+15.27$] & 0\% \\
Boston Housing & Mixed             & $+2.7\%$  & $-0.02\%$ & 4.02 & [$-7.22$, $+6.55$]  & 42\% \\
Cancer         & Prior-Mean        & $+3.9\%$  & $+2.37\%$ & 2.05 & [$-0.30$, $+7.25$]  & 12\% \\
Cancer         & Independent-Prior & $+2.8\%$  & $+1.13\%$ & 1.08 & [$-1.11$, $+3.33$]  & 10\% \\
Cancer         & Correlated-Prior  & $+2.1\%$  & $+0.91\%$ & 1.09 & [$-0.89$, $+3.17$]  & 16\% \\
Cancer         & Mixed             & $+2.7\%$  & $+1.65\%$ & 1.43 & [$-0.20$, $+5.10$]  & 12\% \\
Heart Disease  & Prior-Mean        & $+0.9\%$  & $+0.08\%$ & 0.68 & [$-0.76$, $+1.50$]  & 57\% \\
Heart Disease  & Independent-Prior & $+0.8\%$  & $-0.07\%$ & 0.64 & [$-0.79$, $+1.20$]  & 63\% \\
Heart Disease  & Correlated-Prior  & $+0.9\%$  & $+0.18\%$ & 0.74 & [$-0.61$, $+1.81$]  & 50\% \\
Diabetes       & Prior-Mean        & $+0.8\%$  & $+1.40\%$ & 0.66 & [$+0.26$, $+2.62$]  & 0\% \\
Diabetes       & Independent-Prior & $+1.1\%$  & $+1.78\%$ & 0.77 & [$+0.59$, $+3.15$]  & 0\% \\
Diabetes       & Correlated-Prior  & $+1.0\%$  & $+1.81\%$ & 0.78 & [$+0.59$, $+3.20$]  & 0\% \\
Adult          & all families      & $\approx 0$ & $-0.01\%$ & $\le 0.02$ & [$-0.05$, $+0.00$] & --  \\
\bottomrule
\end{tabular}
\end{table}

\clearpage

\section{Threshold-dependent metrics under label shift}\label{secA:threshold}

We also computed the threshold-dependent metrics, and report them here rather than in the main results because they turn out to measure the decision threshold more than the model. Both classification datasets show large and apparently significant $F_1$ improvements at the conventional $0.5$ threshold: $+0.048$ $[+0.017, +0.080]$ for Prior-Mean on Heart Disease and $+0.089$ $[+0.039, +0.143]$ on Diabetes, significant in all eight results (four prior families on two datasets). These gains are an artefact of the decision threshold, as we now show.

The domain splits induce label shift as well as covariate shift: outcome prevalence moves from $0.212$ in the Diabetes source training pool to $0.495$ in the OOD set, and from $0.459$ to $0.530$ on Heart Disease. A threshold calibrated for one base rate is not appropriate for another \citep{saerens2002adjusting}, so $0.5$ is not a neutral comparison point: any model whose predicted probabilities are shifted upward will appear better on $F_1$ without being a better model. Since the prior-informed models are shrunk toward a non-zero target, they produce more extreme probabilities than the baseline and are closer to the appropriate operating point by accident.

\begin{figure}[htbp]
\centering
\includegraphics[width=\linewidth]{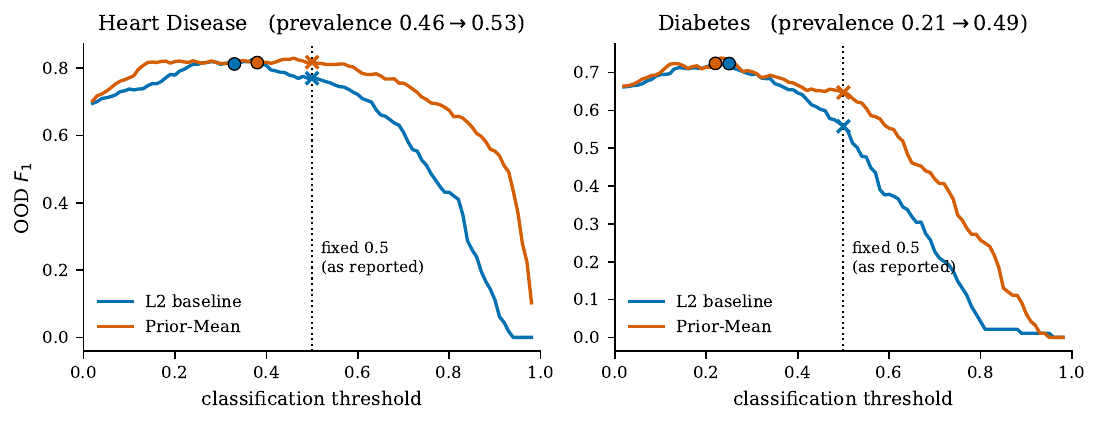}
\caption{$F_1$ on the OOD set as a function of the classification threshold,
for the deployed $L_{2}$ baseline (blue) and the deployed Prior-Mean model
(orange). Crosses mark $F_1$ at the fixed $0.5$ threshold; filled circles mark
$F_1$ at the threshold chosen by 5-fold cross-validation on source data only,
which is the operating point a practitioner would actually deploy. Panel titles
give the prevalence shift from the source training pool to the OOD set. Both
curves peak well below $0.5$ because prevalence rises between the domains, so at
$0.5$ the baseline has already fallen off its peak while the prior-informed
model has not. The prior-informed model's apparent $F_1$ advantage at $0.5$ is
therefore an operating-point effect, and it disappears once the baseline is
given the same threshold freedom.}
\label{fig:threshold}
\end{figure}

Giving every model, baseline included, a threshold chosen by 5-fold CV on source data only removes the effect entirely. Cross-validation selects $0.33$ on Heart Disease, $0.25$ on Diabetes and $0.33$ on Adult: for \emph{every} model, prior-informed or not. At those thresholds all eight previously significant $F_1$ gains lose significance. On Diabetes the recalibrated baseline reaches $F_1 = 0.7236$, above the $0.6472$ that the prior-informed model attains at $0.5$; recalibration alone gains the baseline $+0.166$, nearly twice the $+0.089$ credited to the prior. Independent-Prior ends up behind the recalibrated baseline. Figure~\ref{fig:threshold} shows why: both $F_1$ curves peak well below $0.5$, and at $0.5$ the baseline has already fallen off its peak while the prior-informed model has not.

We report this as a methodological result rather than a footnote. Threshold-dependent metrics at a fixed cut-off are common in this literature, and under label shift they can manufacture an effect where none exists. The threshold-free improvements in log loss and Brier score reported above are unaffected by this critique, and are the defensible form of the classification claim.

\section{Sensitivity to the variance floor}\label{secA:epsilon}

The Independent-Prior penalty weight on coefficient $j$ is
$\lambda/(\widehat{\sigma}^{2}_{j} + \varepsilon)$, so when an elicitation is
unanimous and $\widehat{\sigma}^{2}_{j} = 0$ the floor $\varepsilon$, not the
elicitation, sets how hard that coefficient is held at $\widehat{\mu}_{j}$. One
deployed variant returns bit-identical draws on four of five datasets, so the
question is whether the reported lifts are a property of the elicited spread or
of the floor. Table~\ref{tab:epsilon} re-tunes $\lambda$ and refits at each of
five values of $\varepsilon$ spanning seven orders of magnitude, and for
Correlated-Prior also replaces the jittered inverse with a Ledoit--Wolf
shrinkage estimator of $\widehat{\Sigma}$, which is the principled alternative.

The headline results are not artefacts. Boston Housing's Correlated-Prior lift
is $+11.15\%$ at the deployed setting and $+11.10\%$ under Ledoit--Wolf, and
Cancer is flat to two decimal places across the whole range. Lifts decay only
once $\varepsilon$ reaches $10^{-2}$ or $10^{-1}$, which is where $\varepsilon$ becomes larger than the elicited variances themselves and the Independent-Prior
collapses toward Prior-Mean, so the decay is evidence that the elicited spread contains information, not evidence against it. Heart Disease's Independent-Prior
is the most floor-sensitive case, falling from $+0.75\%$ to $+0.09\%$, and that
is the one place where the reported number should be read with the floor in
mind.

\begin{table}[htbp]
\centering
\footnotesize
\setlength{\tabcolsep}{2pt}
\caption{OOD lift (\%) for the two families whose precision depends on the
elicited spread, as the variance floor $\varepsilon$ is varied and $\lambda$
re-tuned at each setting. ``Zero-spread'' counts features for which the
deployed variant returned bit-identical values across all $M = 50$ draws, so
that $\widehat{\sigma}^{2}_{j} = 0$ exactly. The Ledoit--Wolf column replaces
the jittered pseudo-inverse of $\widehat{\Sigma}$ with a shrinkage estimator
and applies only to Correlated-Prior. The deployed configuration is
$\varepsilon = 10^{-8}$. Lift is MSE reduction for the regression datasets and
AUC gain for the classification ones.}
\label{tab:epsilon}
\begin{tabular}{@{}llcrrrrrr@{}}
\toprule
Dataset & Family & Zero-spread & $10^{-8}$ & $10^{-6}$ & $10^{-4}$ & $10^{-2}$ & $10^{-1}$ & Ledoit--Wolf \\
\midrule
Boston Housing & Independent-Prior & 4/11 & $+10.29$ & $+10.29$ & $+10.30$ & $+9.27$ & $+7.53$ & --  \\
Boston Housing & Correlated-Prior  & 4/11 & $+11.15$ & $+11.15$ & $+11.15$ & $+9.70$ & $+7.69$ & $+11.10$ \\
\midrule
Cancer         & Independent-Prior & 0/13 & $+2.78$ & $+2.78$ & $+2.78$ & $+2.75$ & $+2.44$ & --  \\
Cancer         & Correlated-Prior  & 0/13 & $+2.12$ & $+2.12$ & $+2.12$ & $+2.12$ & $+2.08$ & $+2.20$ \\
\midrule
Heart Disease  & Independent-Prior & 0/18 & $+0.75$ & $+0.75$ & $+0.73$ & $+0.41$ & $+0.09$ & --  \\
Heart Disease  & Correlated-Prior  & 2/18 & $+0.94$ & $+0.95$ & $+0.90$ & $+0.93$ & $+0.93$ & $+0.93$ \\
\midrule
Diabetes       & Independent-Prior & 2/7  & $+1.00$ & $+1.00$ & $+0.90$ & $+0.86$ & $+0.85$ & --  \\
Diabetes       & Correlated-Prior  & 2/7  & $+0.97$ & $+0.97$ & $+0.97$ & $+0.86$ & $+0.84$ & $+0.87$ \\
\midrule
Adult          & Independent-Prior & 0/20 & $-0.03$ & $-0.01$ & $-0.00$ & $+0.00$ & $+0.00$ & --  \\
Adult          & Correlated-Prior  & 0/20 & $-0.01$ & $-0.01$ & $-0.03$ & $+0.00$ & $-0.00$ & $-0.01$ \\
\bottomrule
\end{tabular}
\end{table}

\section{Example elicitation prompt}\label{secA:prompt}

The prompts are assembled from a fixed system part and a dataset-specific user
part, as described in Section~\ref{subsec:elicit}. Both are reproduced verbatim
below for the Diabetes dataset in non-reasoning sampling mode; the other
datasets differ only in the user part. Three non-ASCII characters in the
original ($\mu$, an em dash and $\approx$) are rendered here as \texttt{u},
\texttt{--} and \texttt{\textasciitilde=}, and long lines are wrapped. Note
that the feature-scale summaries are computed on the source training pool only,
and that the domain context describes the source population without revealing
the out-of-distribution target.

\subsection*{System prompt}
\begin{footnotesize}
\begin{verbatim}
You are helping construct informative priors for a regression model.
Adopt the perspective of an endocrinologist and applied statistician with experience
  modelling type 2 diabetes risk in screening data.
Use the feature definitions and feature-scale summaries to reason about BOTH sign and
  magnitude.
All coefficients must be given in the original raw-feature units used by the model
  matrix.
Do not standardise, rescale, or reinterpret the coefficients as standardised effects.
Do not include an intercept.
Output must be only a JSON object matching the provided schema.
Do not include explanations, markdown, code fences, or extra keys.
Return one plausible coefficient vector draw from your belief distribution.
Do not collapse uncertainty into a single average if multiple plausible settings
  exist.
Keep the draw realistic and domain-grounded rather than intentionally noisy.
Prefer interpretable, conservative, domain-plausible coefficients.
Avoid extreme magnitudes unless they are clearly justified by the variable scale and
  domain mechanism.
\end{verbatim}
\end{footnotesize}

\subsection*{User prompt}
\begin{footnotesize}
\begin{verbatim}
Task context
- Binary logistic regression for the positive class in the Pima Indians Diabetes
    outcome (Outcome=1 = patient developed diabetes within 5 years of the visit):
    logit(P(y=1)) = intercept + sum_j beta_j * x_j

Domain context:
- Source domain is younger patients (Age <= dataset median, computed on the
    source-domain subset).
- Priors should encode stable physiology associations likely to transfer beyond the
    source age range.
- Cohort is Pima Indians women aged 21+; the data was collected to study type 2
    diabetes onset.

Dataset notes:
- Age is used only for the OOD split definition and is excluded from predictors.
- Several features (Glucose, BloodPressure, SkinThickness, Insulin, BMI) have observed
    zeros that are likely missing values; the pipeline imputes them with the
    source-domain median.
- DiabetesPedigreeFunction is a derived risk score capturing family history of
    diabetes; it is unitless on a roughly 0-3 scale.

Guidance:
- Use signs and magnitudes consistent with established type 2 diabetes risk-factor
    evidence.
- Prefer conservative values when uncertainty is high.
- Avoid extreme magnitudes unless there is a strong physiological rationale.

Feature meanings:
- Pregnancies: number of times pregnant
- Glucose: plasma glucose concentration after 2-hour oral glucose tolerance test
  (mg/dL)
- BloodPressure: diastolic blood pressure (mm Hg)
- SkinThickness: triceps skin fold thickness (mm)
- Insulin: 2-hour serum insulin (uU/mL)
- BMI: body mass index (weight in kg / height in m^2)
- DiabetesPedigreeFunction: diabetes pedigree function -- a score reflecting family
    history of diabetes

Feature scale summaries from the actual design matrix:
- Pregnancies: continuous/raw numeric; median=2, IQR=[1, 3], central95%~=[0, 5]
- Glucose: continuous/raw numeric; median=110, IQR=[95, 130], central95%~=[78.7, 174],
    missing=0.006
- BloodPressure: continuous/raw numeric; median=68, IQR=[60, 76], central95%~=[50,
    88.1], missing=0.051
- SkinThickness: continuous/raw numeric; median=27, IQR=[19, 35], central95%~=[12.6,
    45], missing=0.203
- Insulin: continuous/raw numeric; median=105, IQR=[66, 168], central95%~=[36, 326],
    missing=0.383
- BMI: continuous/raw numeric; median=31.2, IQR=[25.9, 36.8], central95%~=[21, 44.6],
    missing=0.019
- DiabetesPedigreeFunction: continuous/raw numeric; median=0.369, IQR=[0.241, 0.591],
    central95%~=[0.143, 0.964]

Relevant domain patterns to weigh alongside the summaries:
- Higher Glucose is the strongest single predictor; expect a clear positive
    coefficient.
- Higher BMI raises diabetes risk; the coefficient should be positive but smaller than
    Glucose's.
- Higher Pregnancies and DiabetesPedigreeFunction tend to increase risk modestly.
- BloodPressure, SkinThickness, and Insulin effects are typically smaller and noisier;
    coefficients can be near-zero.

Important modeling reminders:
- Binary indicators use 0/1 coding, so their coefficients represent the change
    associated with turning the indicator on.
- Continuous-feature coefficients should be scaled to the feature summaries shown
    above; do not assume a one-unit change is small unless the summary suggests it is.
- When multiple variables capture overlapping structure, distribute signal
    conservatively rather than giving each one a large effect.
- The output schema already fixes the keys. Choose only the numeric values.
\end{verbatim}
\end{footnotesize}

\end{appendices}

\end{document}